\documentclass[preprint]{elsarticle}

\usepackage{graphicx}
\usepackage{lineno,hyperref}
\usepackage{booktabs,tabu}
\usepackage[singlelinecheck=false,labelfont=bf]{caption}
\usepackage[flushleft]{threeparttable}
\modulolinenumbers[5]
\makeatletter
\def\ps@pprintTitle{%
 \let\@oddhead\@oddhead
 \let\@evenhead\@empty
 \def\@oddfoot{\leftline{Accepted in Applied Soft Computing}}%
 \let\@evenfoot\@oddfoot}
\makeatother

\journal{Applied Soft Computing}
\begin{document}

\begin{frontmatter}

\title{Estimation of lactate threshold with machine learning techniques in recreational runners}

\author[mainaddress]{Urtats Etxegarai\corref{cor1}}
 \ead{urtats.etxegarai@ehu.eus}
 \cortext[cor1]{Corresponding author.Tel.:+34 94 601 3910}
\author[mainaddress]{Eva Portillo}
\author[addressTwo]{Jon Irazusta}
\author[mainaddress]{Ander Arriandiaga}
\author[mainaddress]{Itziar Cabanes}

\address[mainaddress]{Department of Automatic Control and System Engineering, Faculty of Engineering, University of the Basque Country UPV/EHU, Bilbao, Spain}
\address[addressTwo]{Department of Physiology, Faculty of Medicine and Nursing, University of the Basque Country UPV/EHU, Leioa, Spain}

\begin{abstract}

Lactate threshold is considered an essential parameter when assessing performance of elite and recreational runners and prescribing training intensities in endurance sports. However, the measurement of blood lactate concentration requires expensive equipment and the extraction of blood samples, which are inconvenient for frequent monitoring. Furthermore, most recreational runners do not have access to routine assessment of their physical fitness by the aforementioned equipment so they are not able to calculate the lactate threshold without resorting to an expensive and specialized center. Therefore, the main objective of this study is to create an intelligent system capable of estimating the lactate threshold of recreational athletes participating in endurance running sports.

The solution here proposed is based on a machine learning system which models the lactate evolution using recurrent neural networks and includes the proposal of standardization of the temporal axis as well as a modification of the stratified sampling method. The results show that the proposed system accurately estimates the lactate threshold of 89.52\% of the athletes and its correlation with the experimentally measured lactate threshold is very high (R=0,89). Moreover, its behaviour with the test dataset is as good as with the training set, meaning that the generalization power of the model is high. 

Therefore, in this study a machine learning based system is proposed as alternative to the traditional invasive lactate threshold measurement tests for recreational runners.
\end{abstract}

\begin{keyword}
\texttt{Lactate}\sep Anaerobic Threshold \sep Modelling \sep Evolution \sep Machine Learning \sep Neural Network\\
\hspace{20 mm}\\
 $\copyright$ 2018. Available under the CC-BY-NC-ND 4.0 license http://creativecommons.org/licenses/by-nc-nd/4.0/
\end{keyword}

\end{frontmatter}

\section{Introduction}
\label{sec:intro}

In recent years, the popularity of endurance sports has dramatically increased, especially in long distance running events and triathlon. During 2015, in the USA alone, 17 million people finished a running event \cite{Runningusa} and more than 4 million people participated in triathlons \cite{Triathlons}. Among this heterogeneous population, a significant amount of athletes train methodologically to improve their performance \cite{Vickers2016}.

Understanding the energy supply mechanisms of the human body is of great interest for sports performance. The human body uses different energy supply systems depending on the intensity and duration of the activity performed. In long duration exercises as endurance sports, the oxidative or aerobic energy system is the main energy contributor as more powerful anaerobic systems are not sustainable in the long term without creating excessive fatigue. Therefore, the transition between the predominant use of aerobic to anaerobic energy supply systems, i.e. anaerobic threshold, plays a key role in the performance of the athlete \cite{Tanaka1984a,Pallares2016}.

Lactate threshold is the exercise intensity at which the concentration of blood lactate begins to significantly increase compared to the values at resting and is closely related with the anaerobic threshold. This is considered an essential parameter when assessing performance of elite and recreational runners \cite{Pallares2016,Lacour1990}, prescribing training intensities in endurance sports\cite{Acevedo1989a,Hofmann2017} and can be useful for athlete internal load monitoring \cite{Halson2014}. Furthermore, nowadays it is fully demonstrated that the lactate threshold is more decisive for endurance sports performance than other variables such as the maximum consumption of oxygen or the running economy \cite{Tanaka1984,Heck1985,Fay2009,Grant1997}. 

However, the measurement of blood lactate concentration requires expensive equipment and the extraction of blood samples, which are inconvenient for frequent monitoring. Furthermore, most recreational runners do not have access to routine assessment of their physical fitness by the aforementioned equipment so they are not able to calculate the lactate threshold without resorting to an expensive and specialized centre. Therefore, the main objective of this study is to create an intelligent system capable of estimating the lactate threshold of recreational athletes participating in endurance running sports. In order to attain a model applicable to the real world, a non-invasive, cost efficient, dependant upon easily measurable features and easily accessible solution is essential.

The complexity of this problem lies on the non-linear dynamic behaviour of lactate production and the multiple variables involved. This dynamic behavior is well characterized by the individual lactate threshold calculated with the Dmax method and nowadays stands as the most recommended technique \cite{Santos-Concejero2014a}. In this method, each athlete performs an incremental treadmill speed test to calculate its corresponding individual lactate curve so that the lactate threshold can be calculated from it using the Dmax method.

\begin{figure}
\includegraphics[width=\columnwidth]{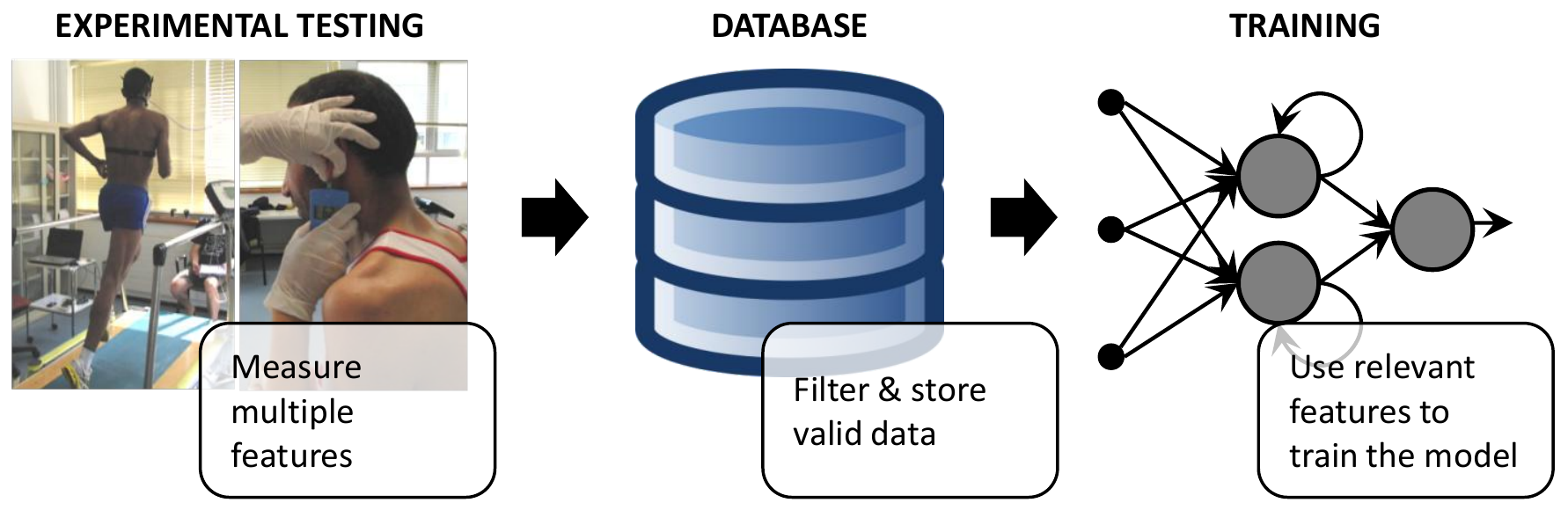}
\caption{Lactate production modelling process}
\label{fig:model}
\end{figure}

Thus, the methodology here proposed models the dynamic of the lactate production in the aforementioned incremental treadmill speed test conditions. More precisely, Recurrent Neural Networks (RNN) have been used to learn the non-linear and dynamic behavior of lactate metabolism and tie it with several easily measurable physiological features assessed during these tests. Moreover, as represented in figure \ref{fig:model}, an extensive experimental database obtained through running tests is used to train the model. These tests were performed to a wide variety of endurance recreational athletes thus covering a wide range of this heterogeneous population.

\begin{figure}
\includegraphics[width=\columnwidth]{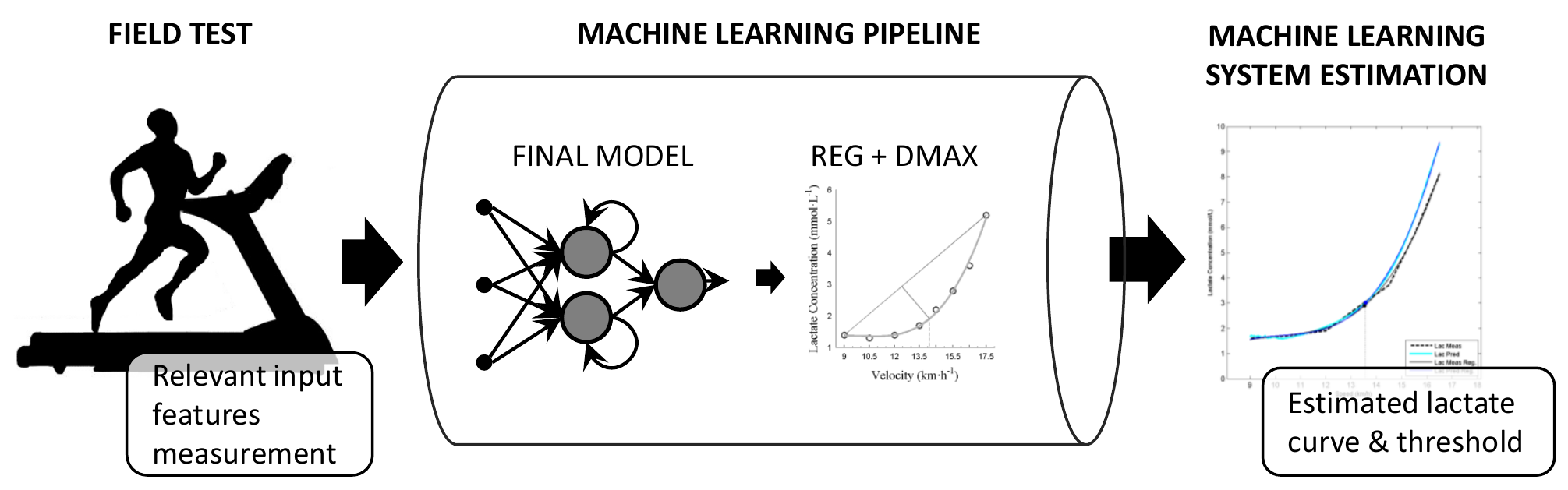}
\caption*{Abbreviations: Reg, regression}
\caption{Lactate threshold estimation procedure}
\label{fig:LTestimation}
\end{figure}

This way, the machine learning (ML) based system here presented is used as an alternative to the traditional invasive lactate threshold measurement test. As represented in figure \ref{fig:LTestimation}, the athlete interested in estimating its lactate threshold performs the aforementioned running speed test but only assessing the easily measurable input features. These input features are then introduced in the ML system (pipeline) which estimates the individual lactate threshold of the athlete.

The rest of the article is organized as follows: Section \ref{sec:background} reviews the state-of-art analyzing the lactate threshold estimation approaches proposed so far; section \ref{sec:strategy} presents the methodology followed in the design of the ML system; section \ref{sec:experimental} describes the experimental setup; section \ref{sec:results} presents the results and discussion; lastly section \ref{sec:conclusions} offers concluding remarks.

\section{Background and related work}
\label{sec:background}

\begin{figure}
\includegraphics[width=\columnwidth]{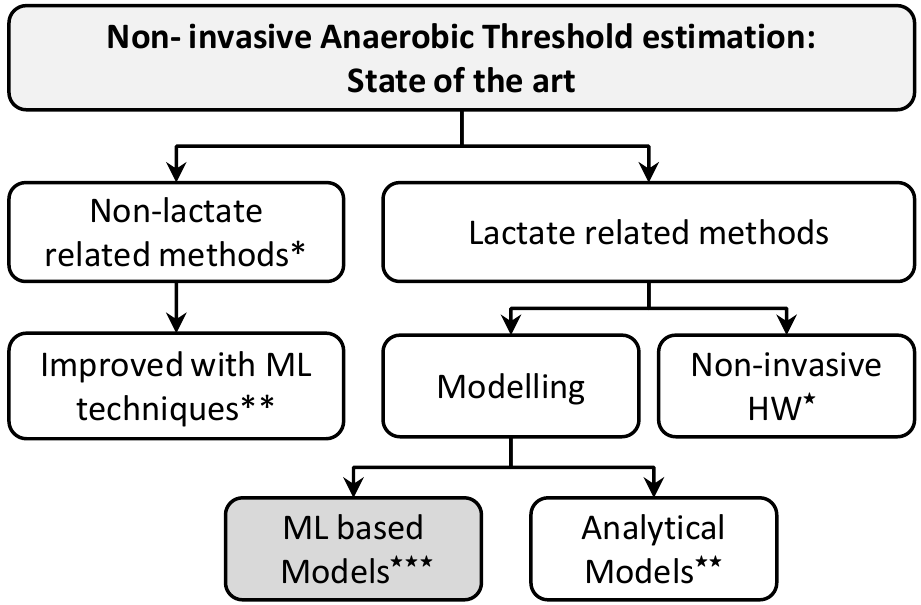}
\caption*{Abbreviations: HW, hardware; ML, machine learning}
\caption*{$*$ \protect\cite{Peinado2016,Conconi1982,Vachon1999,Bourgois2004};
				$*$ $*$ \cite{Ringwood2014};
				$\star$ \cite{Borges2015};
				$\star$$\star$ \cite{Proshin2013};
				$\star$$\star$$\star$ \cite{Erdogan2009}.}
\caption[]{Non-invasive anaerobic threshold estimation: State of the art}
\label{fig:stateofartSC}
\end{figure}

Traditionally, the anaerobic threshold estimation has relied on taking direct blood lactate samples. However, great efforts have been made in order to find a non-invasive solution to the anaerobic threshold estimation. As it is shown in figure \ref{fig:stateofartSC}, very diverse approaches have been proposed in the literature. 

Among these approaches, some studies tried to estimate the anaerobic threshold using other physiological parameters different to lactate. One of the most extended methods is based on the measurement of expired gases \cite{Peinado2016}. This technique has shown to be an accurate solution when assessing the anaerobic threshold. Nevertheless, it still uses expensive and specialized equipment to make estimations and therefore, its not valid to fulfill our objectives. On the other hand, heart rate deflection point (HRDP) has also been used as indicator of the anaerobic threshold. The traditional study that put this relation into practice is known as Conconi Test \cite{Conconi1982}. Since then, several other studies followed this approach while some others point out its deficiencies \cite{Vachon1999,Bourgois2004}. These studies evidenced that, in spite of being a commonly used method, this method is far from being infallible. Factors such as lack of control of the conditions during and previous to the test and the impossibility of finding the HRDP in all the cases are probably among the reasons which make this method inaccurate. 

In this regard, a ML application has also been used to improve the estimation of the HRDP and thus solving one of the weak points of the methodology originally proposed by Conconi \cite{Ringwood2014}. More precisely, Ringwood et al. proposed to use non-linear autoregresive exogenous (NARX) models in combination with fuzzy interpolation to model the heart rate (HR) dynamics and to find the HRDP. However, the accuracy obtained was low and the population studied very small (9 athletes) and therefore its not reasonable to draw conclusions about its applicability, specially about the generalization capabilities of the model.

Other strategies have focused on using the lactate as indicator of the anaerobic threshold but avoiding its direct measurement.

The use of non-invasive lactate measurement devices such as the one presented by Borges et. al. \cite{Borges2015} is among this type of strategies. However, despite having good accuracy and not taking direct blood samples, the devices are still expensive and can cause extra discomfort specially in sports such as triathlon were the transitions between disciplines are determinant.

It is known that the blood lactate concentration is related with multiple features of the athlete such as the HR at a given speed (or the speed at a given HR), the rate of reduction of the HR after an exercise or heart rate recovery, the rate of perceived exertion (RPE) as described by Borg \cite{Borg1982a}, gender, age, diet or athlete level \cite{LopezChicharro2004}. Nevertheless, nowadays there is no analytic equation that solves this complex non-linear and dynamic problem and finding an analytical model with these characteristics for a complex system as the human body, is far from been feasible. 

Actually, Proshin et al. \cite{Proshin2013} proposed a mathematical model of human lactate metabolism gathering several physiological models and merging them into a single one. This model was then parametrized to fit to individual athletes. However, the authors acknowledged that this approach was not valid for all type of athletes, being more appropriate for high-speed sports. Moreover, the parametrization of the model required from multiple measurements including blood lactate measurements and a respiratory metabolism analysis among others. This leaves this study out of the scope of this work as it does not fulfill the objectives previously presented. In any case, this study highlights the complexity of the lactate modelling problem even if individualization is the only concern. The complexity increases if both accuracy and generalization are sought, as it is our case. 

In complex problems such as human lactate metabolism where the multiple factors underlying the lactate production are still not clear, empirical or measurement oriented modelling strategies are probably more suitable. In this regard, Artificial Neural Networks (ANN) are widely used to create empirical models of non-linear dynamic problems \cite{Zhang2011,Godarzi2014,Becerikli2007,Babu2014,Laboissiere2015}. Actually, they have also been scarcely used to model lactate production in athletes. In particular, Erdogan et. al. \cite{Erdogan2009} proposed a model based on a multi-layer perceptron (MLP) to estimate the HR at onset of blood lactate accumulation (OBLA) point, which corresponds to a fixed blood lactate concentration of 4 mmol/l. However, it is known that the OBLA point, as it corresponds to a fixed blood lactate concentration, does not take into account the inter-individual variability of LT, specially in runners of varying athletic ability as in our case. This is why the individual lactate threshold calculated by the Dmax method is currently the most recommended methodology \cite{Santos-Concejero2014a} and the one used in our work. Moreover, since both accuracy-generalization are a non-separable duo (closely related to bias-variance dilemma) \cite{Dreyfus2005} an exclusive focus on accuracy directly affects in a decrease of generalization (and vice versa). Moreover, when modelling complex systems as the human body, obtaining generalization is a much more ambitious challenge than obtaining accuracy and both are essential for the applicability of the model in the real world. Therefore, the validity of the model proposed by Erdogan et. al. \cite{Erdogan2009} is limited by their methodology which exclusively focus on maximizing the accuracy and then estimating their model generalization with their test set. Since the estimated generalization of any model is necessarily limited and cannot extend beyond the boundaries of the region of input space \cite{Dreyfus2005}, the applicability of any model is limited to the information characterized in the database. This problem is exacerbated when the focus is placed on maximizing the accuracy and this approach is more prone to create overfitted models. In any case, this demonstrates that, despite being far from achieving the accuracy and specially the generalization needed to use the models in practice, ML techniques are valid to create models to estimate the lactate threshold.

Usually the dynamic of the output variable itself intrinsically contains data missing in the available input features. This is very common in real time dynamic problems \cite{Zhang2011,Godarzi2014,Becerikli2007,Babu2014,Laboissiere2015}, specially  when the output variable is dependent on multiple not easily measurable or unknown features as in our case. In ML context, this type of problems are known as regression problems, where the parameters of the model are directly learnt from experiments using supervised learning.  

\begin{figure}
\includegraphics[width=\columnwidth]{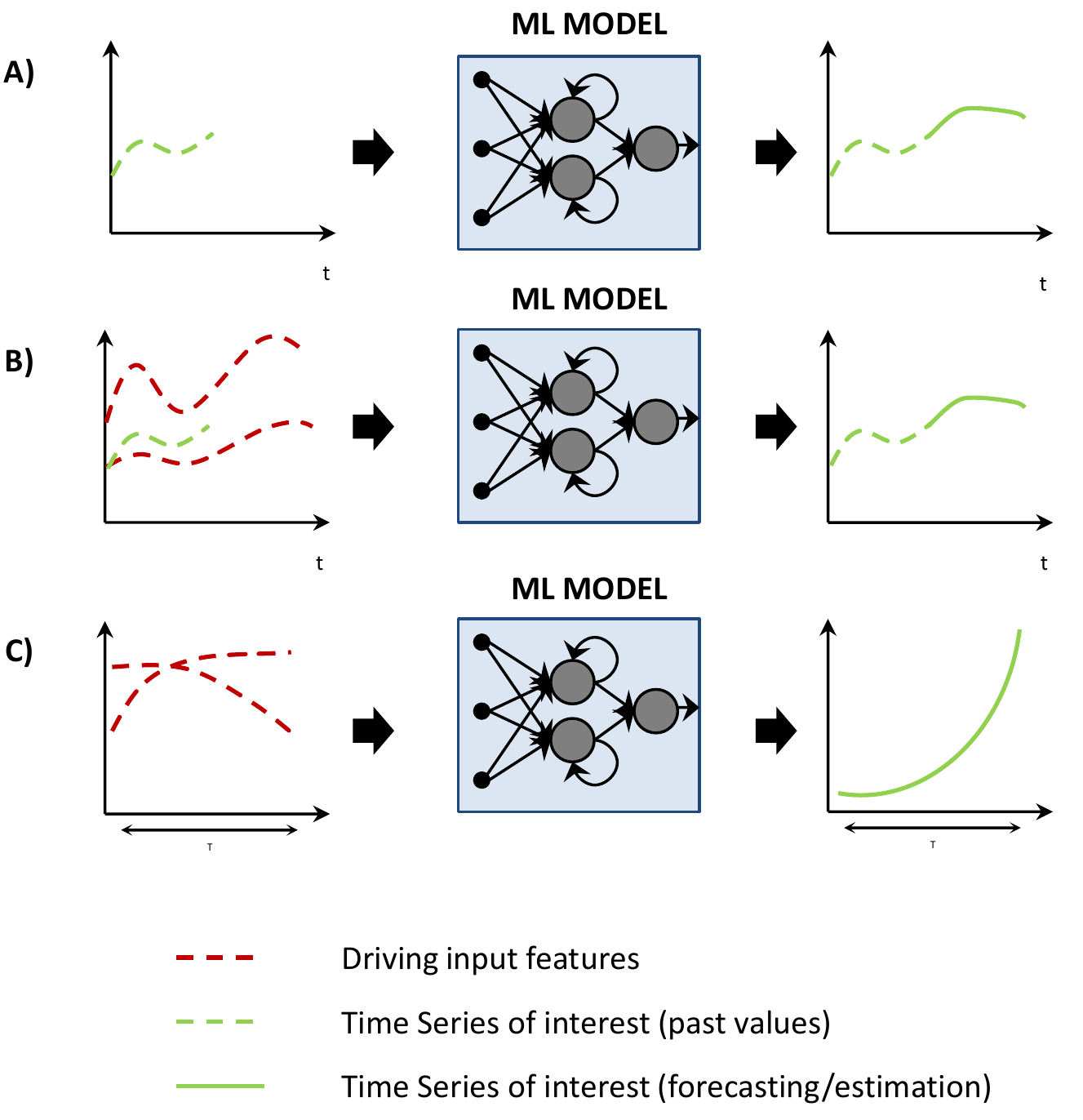}
\caption*{Abbreviations: ML, machine learning}
\caption{Dynamic modelling approaches}
\label{fig:DynamicModelling}
\end{figure}

Most of the dynamic modelling problems focus on predicting future values of a single time series based on current and past values, or in other words, learning from historical data. These problems are also known as time-series prediction problems. Depending on the complexity of the problem, these systems can be modeled either using pure auto-regressive techniques as represented in figure \ref{fig:DynamicModelling}A or enriching it with additional input features figure \ref{fig:DynamicModelling}B \cite{Zhang2011,Babu2014}.

However, there is another research area less extended in the literature due to the particular characteristics of the problem that tries to solve. It focuses on modelling the dynamic behavior of a system in certain specific conditions in which there are no real current and/or past values for forecasting the future ones. This means that the time-series of interest has to be completely estimated from the driving features as is represented in figure \ref{fig:DynamicModelling}C.

For instance, in \cite{Asgari2016} a RNN was used to model dynamic non-linear systems of a gas turbine for simulation of its start-up operation. A NARX model without current time step data was used in this case and the final models were validated against other three experimental data sets. In this work, a single gas turbine was modeled using a database which covered all the start-up situations in which the gas turbines could work.

However, the complexity of this type of problems increases together with the variability of the time-series of interest. This is the case of the lactate metabolism which, due to the complexity and multifactorial nature of the human body, varies a lot from one individual to another, even within endurance runners population. In this type of problems, the key factor determining the successful estimation of the time-series of interest is the capacity of the ML system to explain this variability, or in other words, the generalization capabilities of the model.

This problem is well represented by Teufel et. al. \cite{Teufel2003}. The objective of this study was to model the glucose metabolism of a diabetic person using an Elman NN and thus to avoid taking continuous blood samples. Authors assumed that the short term relation between the blood glucose levels and the combination of ingested carbohydrates and injected insulin stays stable for a particular person. An individual model was trained for each person and the final model was capable of estimating a complete time-series with good accuracy. However, despite the good results obtained for a particular person, the authors noticed that their solution lacks of generalization, since models where not applicable to other people. Probably, one of the main reasons for the bad generalization capabilities of the model is the lack of input data which could explain the inter-individual variability of the glucose metabolism. In this regard, Arriandiaga et al. \cite{Arriandiaga2015a} proposed an approach to deal with this type of problems. They proposed a novel methodology to estimate complete time series in a specific time interval from multiple and distinct input time-series. This is applicable to problems where as in our case, the relation between individuals are deep and the variability too high to be learnt from few experiments.

\section{Dynamic non-linear modelling strategy}
\label{sec:strategy}
\begin{figure}
\includegraphics[width=\columnwidth]{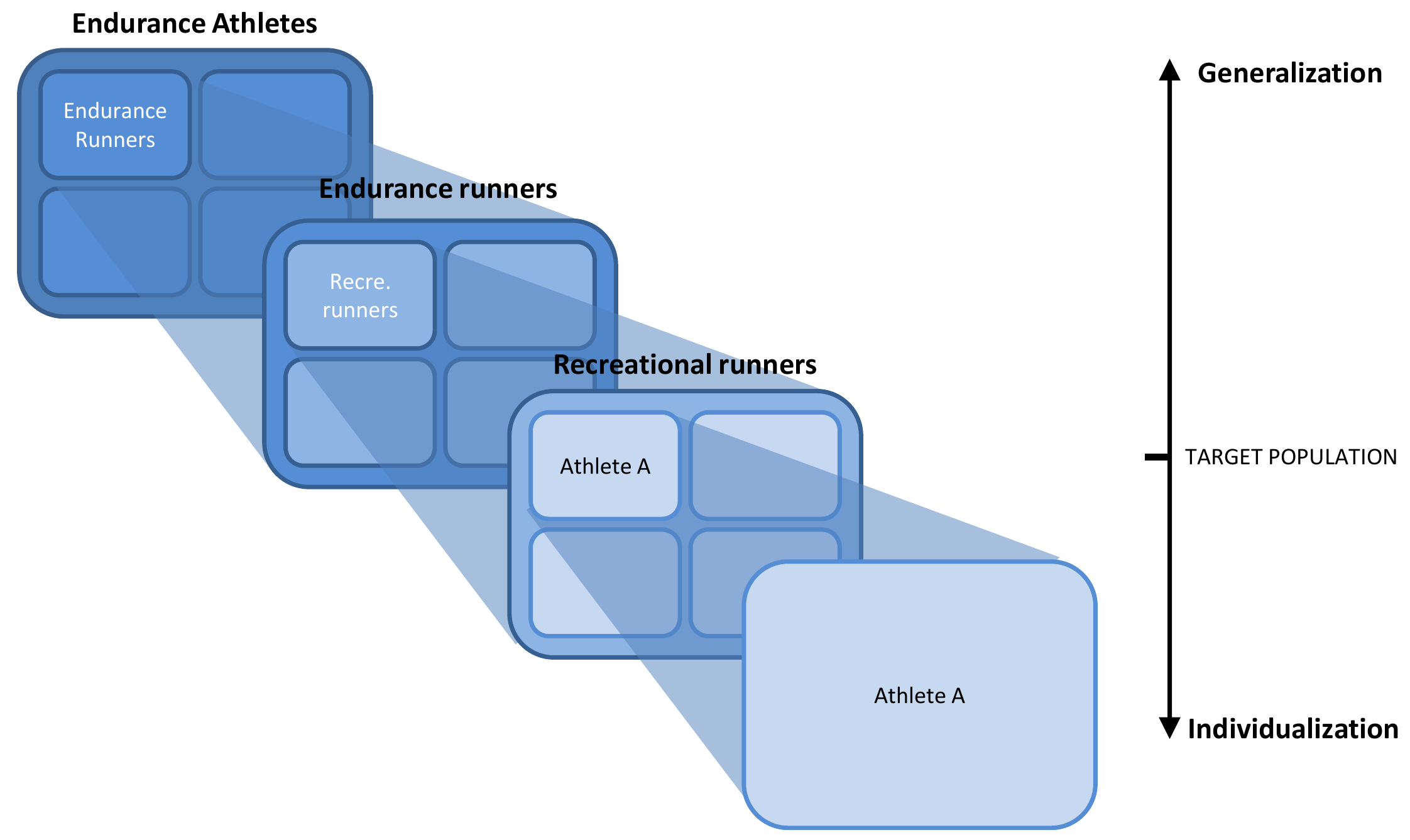}
\caption{Individualization-generalization continuum}
\label{fig:ZoomLevels}
\end{figure}

In order to create a ML system applicable in sports performance while fulfilling the previously established objectives, is essential to build a model around two confronting characteristics. On the one hand, individualization is key for sport performance. On the other hand, a model valid for an entire heterogeneous population is sought. Therefore, as both individual and global characteristics are to be learnt, this problem can be interpreted from a individualization-generalization perspective. Figure \ref{fig:ZoomLevels} illustrates this idea.

\begin{figure}
\includegraphics{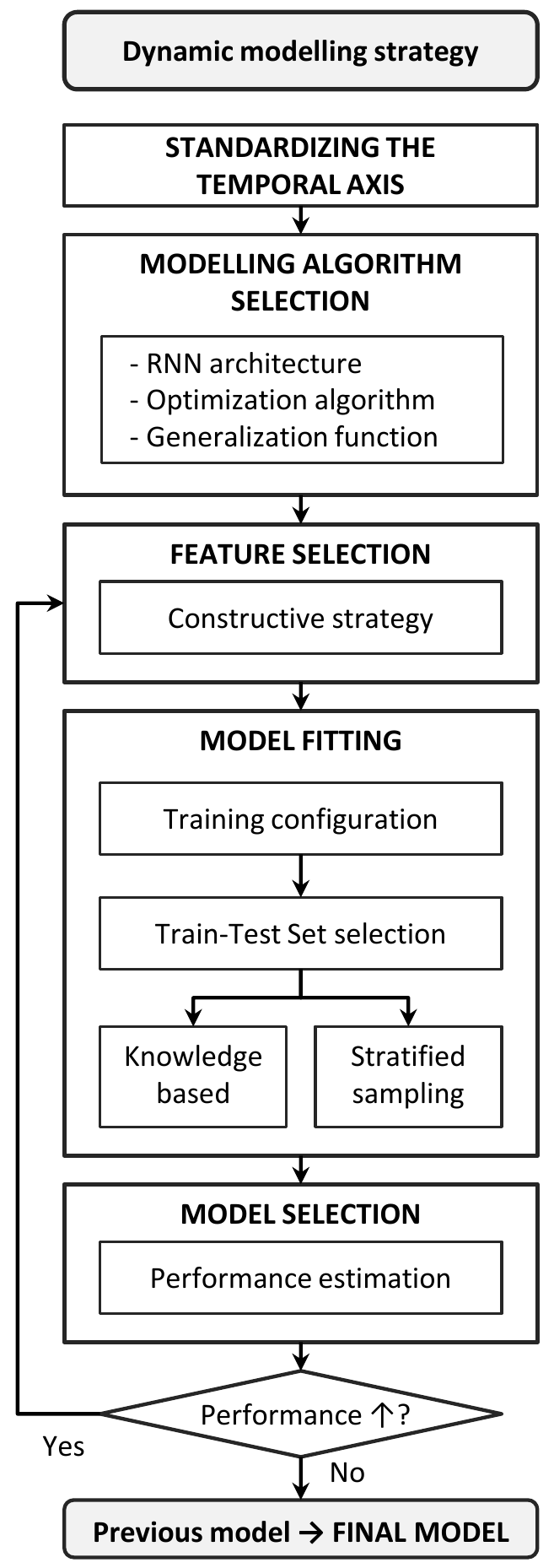}
\caption{Dynamic modelling strategy}
\label{fig:DinamicModellingStrategy}
\end{figure}

Finding a model with both perfect individualization and generalization capabilities would require both a huge database and an extremely powerful algorithm capable of learning very deep relationships. However, creating a database capable of completely characterize the target population is not viable from the economical and technical perspective. Therefore, the design efforts are directed to find a model with the best trade-off between individualization and generalization. To do so, the methodology here proposed creates multiple models covering a wide range of the individualization-generalization continuum. Then, the best model is selected according to ad-hoc created performance indicators. This methodology follows five steps represented in figure \ref{fig:DinamicModellingStrategy} and described in the next sections.

\subsection{Standardizing the temporal axis: Relativization of the exercise intensity}
\begin{figure}
\includegraphics[width=\columnwidth]{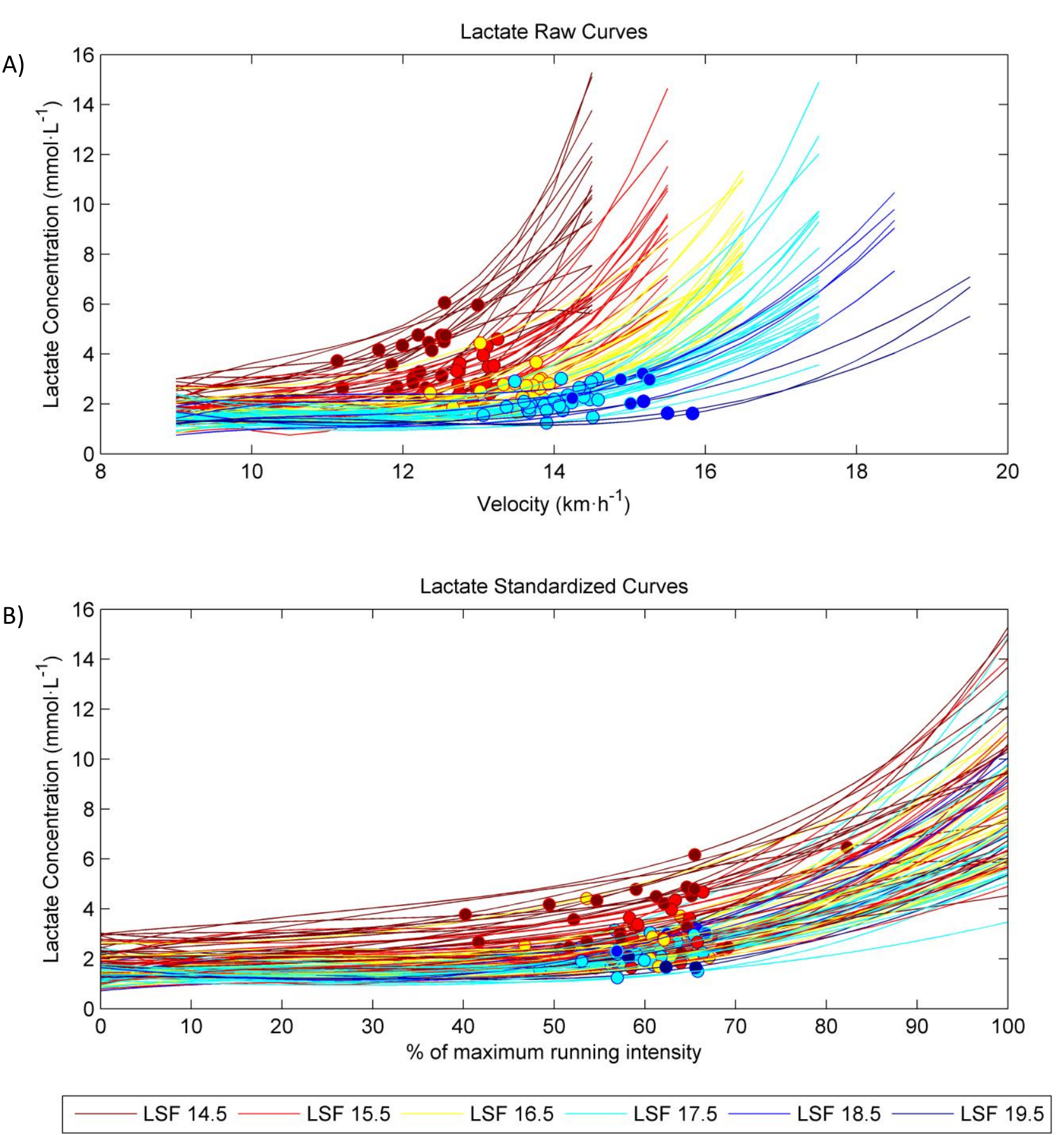}
\caption*{Abbreviations: LSF, last step finished}
\caption{Experimental lactate curves: \\\hspace*{3em}A) raw lactate curves. \\\hspace*{3em}B) relativized lactate curves}
\label{fig:LactateCurves}
\end{figure}

As explained in section \ref{sec:experimental}, the testing protocol undertaken by the athletes is a maximal test which ends when the athlete can no longer maintain the running speed. As more fitted athletes can maintain higher running speeds, the duration of each test depends on the individual fitness level of each athlete. In this case, as recreational athletes have very diverse levels, the tests are also very diverse in length as it is shown in figure \ref{fig:LactateCurves}A. 

Therefore, the time series representing the evolution of the input and output features differ in length. If these raw time-series were used to train the RNN it is presumable that the longer ones would have more relevance in the learning process due to their greater number of data points and meaning that the model would not be valid for athletes with lower peak treadmill speed (PTS). Hence, in this work a novel methodology is proposed to deal with this type of problems.

Due to the particular characteristics of the lactate production while performing an incremental speed test, its evolution has a semi-exponential shape as it is shown in figure \ref{fig:LactateCurves}. Despite the PTS differs from one athlete to another, the curve shape remains similar. Moreover, the individual lactate threshold is not related with an absolute exercise intensity but with the tipping point of the curve. Being this so, in this work a standardization of the temporal axis (i.e. the running speed axis) is proposed for two purposes:
\begin{itemize}  
\item Unify the lengths of all time-series so they have equal relevance in the learning process
\item Concentrate the lactate threshold of all athletes in the same region so that the learning process is simplified
\end{itemize}

By doing that, as the running speed is directly related with the exercise intensity, each test features are relativized with respect to the maximum intensity of each athlete. As shown in figure \ref{fig:LactateCurves}B, the standardization allows to group athletes with different fitness levels by their \% of maximum intensity and treat them as a single group. This approach is useful when as in our case, the dynamic of the system is much more relevant for the final result than the absolute values related to it. 

\subsection{Modelling algorithm selection}
In this work, our approach is not to propose novel modelling algorithms but to investigate the applicability of previously consolidated ones and propose a methodology to solve complex real problems.

\begin{figure}
\includegraphics[width=\columnwidth]{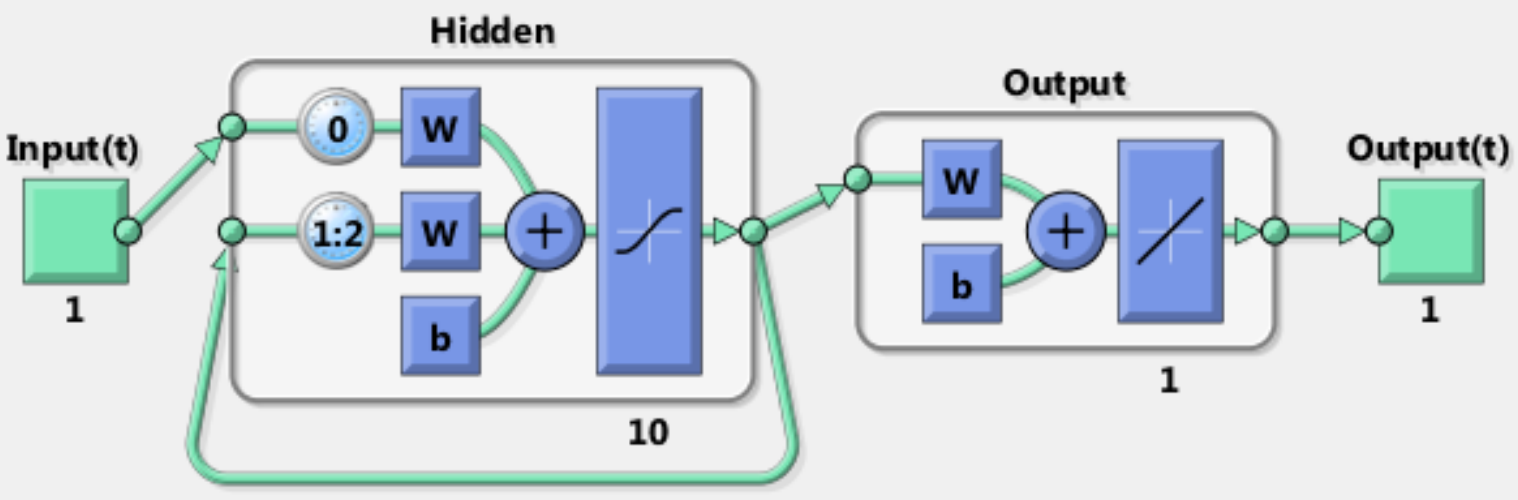}
\caption{Layer-recurrent neural network. Reprinted from \cite{Matlab}} 
\label{fig:LRRN}
\end{figure}

In this case, as represented in figure \ref{fig:LRRN}, a layer-recurrent neural network (LRNN) architecture has been selected. The LRNN is an Elman-inspired recurrent neural network which has flexibility to configure the number of hidden layers and the transfer function of each layer \cite{Arriandiaga2015a}.

In addition, the Levenberg-Marquardt minimization function is used to fit the model parameters as it has a faster convergence and lower error rate than other widely used minimization algorithms \cite{Hagan1994}. 

Finally, the most used generalization methods are the cross-validation and Bayesian regularization. The former divides the database into training, validation and test sets to asses the generalization power of the model. However, Bayesian regularization does not need the validation phase so the database is split only into two sets, for training and for testing. This means that with this approach more experiments can be used for training purposes which allows to take most advantage from the database. In addition, the achieved generalization capabilities can be higher than with cross-validation \cite{MacKay1992,Mackay1995}.

\subsection{Feature Selection}
As previously stated, the inputs of the model shall be easily measurable and contain inter-individual characteristics on it. Among the several physiological features known to be related to blood lactate concentration, features such as HR-derived ones, RPE, age and gender fulfill both criteria.
 
However, the selection of the features to be used as the input of the model is not straightforward. It is known that, for a given accuracy, the model with less parameters, i.e. the most parsimonious, generalizes best \cite{Dreyfus2005}. As the complexity of the model increases together with its number of input features, minimizing the number of inputs is important to obtain a model with better generalization capabilities. Therefore, to select the input features of the model from the candidate inputs, a constructive strategy has been followed. This means that, the simplest model (i.e. a model  with zero input features) is compared to a model with the next most relevant input feature in order to select the best one. This procedure is then repeated until the addition of a new input no longer improves the quality of the previous model. 

The features selected for the constructive strategy are in relevance order:
\begin{enumerate}
\item Evolution of heart rate at the end of each stage (HRevo)
\item Evolution of heart rate after 1 minute rest (HRRevo)
\item Evolution of RPE (RPEevo)
\end{enumerate}

\subsection{Model fitting}
The model has been trained using Matlab R2013b software Neural Network and Statistics and Machine Learning Toolboxes (MathWorks, Natick, Massachusetts, USA).

\subsubsection{Training algorithm configuration}
The training algorithm is configured with two parameters:
\begin{itemize}
\item Hidden Units (HU)
\item Delays 
\end{itemize}

The complexity of each model depends on these training algorithm configuration parameters.

In order to minimize the training process time, is desirable to minimize the range of configuration parameters while ensuring that the final model falls within that range. To do so, a preliminary training analysis is performed using a small portion of the database so that a preliminary range of parameters can be selected at no high computational cost. This preliminary training analysis has three steps:

\begin{enumerate}
\item Coarse tuning: Several trainings are performed with a wide range of parameters trying to cover a big range of model complexities. The performance of these first models is calculated. 
\item Increased resolution on operation point: The configuration parameter ranges are reduced and the resolution increased in order to focus in the zone where the best results have been obtained in the first step. Several models are trained according to this approach and their performances calculated.
\item Fine tuning: If possible, the range is further reduced according to the results obtained in the second step.
\end{enumerate}

Then, the training process is repeated for the complete database with the range of configuration parameters determined in the preliminary analysis. Finally, a sensitivity analysis of the training process is made to deduce if a more detailed configuration parameters selection is needed. 

On the other hand, the function minimum found by the training algorithm depends also on the initial weight values of the model. Therefore, in this work each neural network configuration (with certain HUs and Delays) is trained ten times with different weight initializations. In this case, the algorithm proposed by Nguyen and Widrow \cite{Nguyen1990} has been applied because it reduces training time over other weight initialization methods such as the layer-by-layer or purely random initialization.

\subsubsection{Train and test set selection}

The split of the database into a training and a test set shall be done in a way that the whole target population (i.e. recreational runners) is characterized in both sets.

In this case, a 70-30\% split has been selected for the training and testing sets respectively. Usually, the training and test examples selection is done by random sampling, however, this approach is suitable only if the available database is big enough to make the random selection statistically significant. Therefore, in this work, two other different approaches have been followed and compared.

\paragraph{Knowledge based sampling}
In this approach, the selection of the train and test examples has been made based on the knowledge of the physiology experts. The training and test examples have been selected taking into account the diversity of lactate curve shapes so that all type of shapes were present both in the training and test sets. In this regard, several characteristics of the curves such as the maximum/minimum lactate values, last step reached, decreasing/increasing patterns, non-exponential shapes and other anomalies have been considered.

\paragraph{Modified stratified sampling: Time-series clustering for stratification}
The other sampling approach proposed in this work is a variation of a stratified random sampling. In this method, the whole population is classified into mutually exclusive and more homogeneous groups called strata. Then, a simple random sampling is made from each stratum so that a heterogeneous sample is created containing examples of all the sub-populations.

\begin{figure}
\includegraphics[width=\columnwidth]{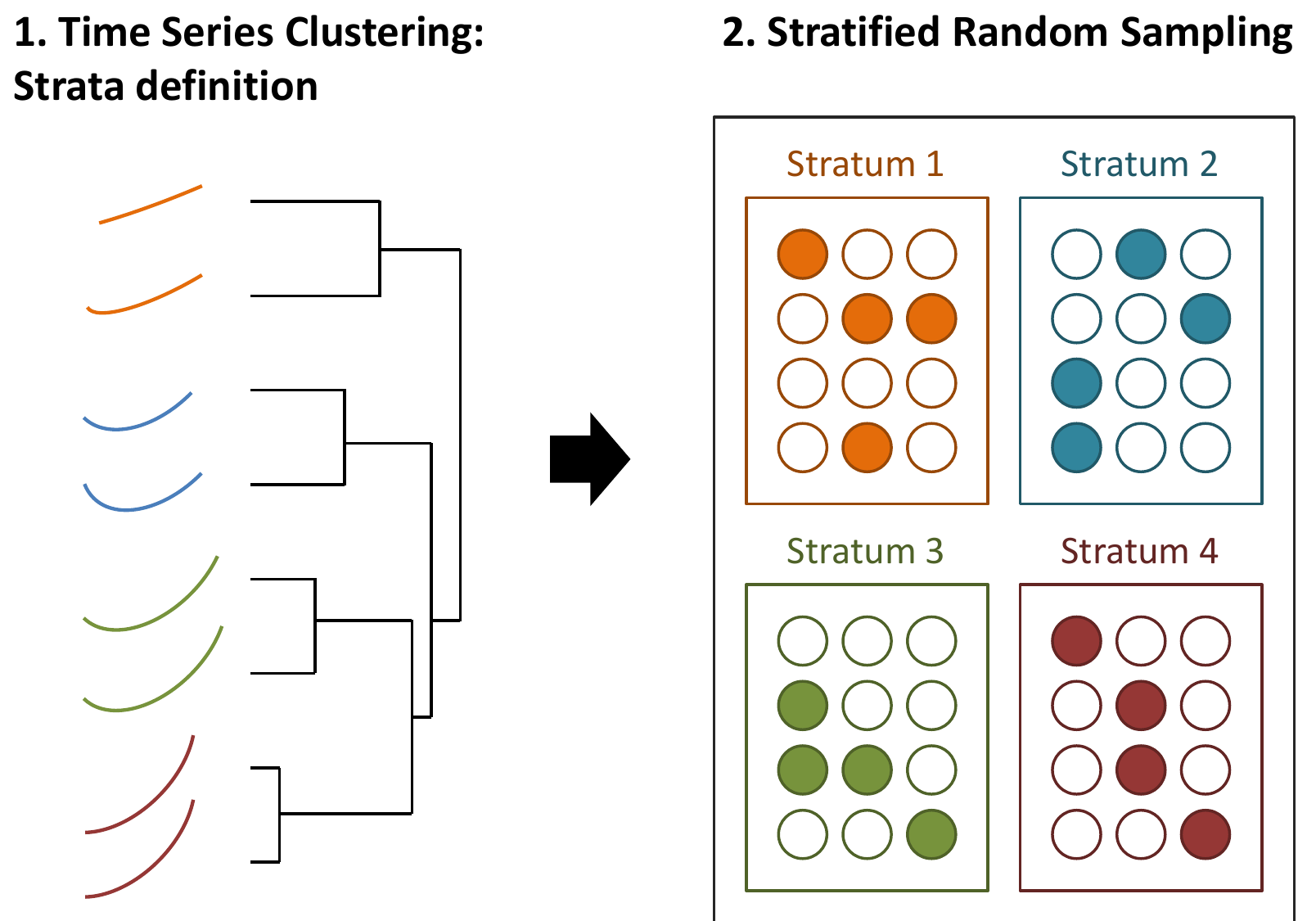}
\caption{Stratified Random Sampling}
\label{fig:SRS}
\end{figure}

Usually, the classification of the whole population in several stratum is made according to one or more static parameters. However, in our case the parameter which is considered most determinant to create the strata is not a static parameter but the lactate curve shape. Therefore, a time-series clustering technique is proposed to make the stratification according to the lactate curve shapes. Figure \ref{fig:SRS} illustrates this methodology.

In this case, a hierarchical time-series clustering algorithm is used to create the strata as it allows not to preselect the number of clusters and to make a point to point curve comparison, which is very convenient in this case where the time-series are standardized to be equal in length. 

Once the strata is defined, the 30\% of the examples of each stratum are randomly selected and included in the test set and the rest in the training set. 

\subsection{Model selection: Ad-hoc performance estimation} 
As previously mentioned, the best trade off between individualization and generalization is sought for the final model.

Accuracy and generalization, as defined in this work, are closely related to what in ML is known as bias-variance dilemma \cite{Ng2012}. Since generalization is a non-computable value and can only be estimated according to the boundaries of your input space \cite{Dreyfus2005}, our work proposes a methodology focused on maximizing the generalization. First, a 'minimum necessary accuracy' is set and then the simplest model meeting this accuracy is selected as the best model. This methodology is based on the 'principle of parsimony' which is the best way to maximize the generalization power of the model \cite{Dreyfus2005}. Although this methodology does not ensure that the generalization power of the final model is high (or even good enough), it does ensure that the generalization power of the model is maximized for the boundaries of your particular problem.  

\subsubsection{Model performance estimation: Approaching theory and practice}
In this work, two different model performance indicators (statistical and heuristic) are used to determine the overall performance of the models so that both theoretical and practical perspectives are taken into account when selecting the final model.

A statistical indicator is used to asses the performance of the model calculating the correlation between the real individual lactate threshold and the estimated individual lactate threshold. The real individual lactate threshold is considered the one calculated with the lactate levels measured in the running test while the estimated individual lactate threshold is determined by the ML system here proposed. Both lactate thresholds are calculated using the Dmax method of each corresponding curves as it is recommended for recreational athletes \cite{Machado2012,Nicholson2001}. 

However, this approach does not help to determine the applicability of the model to sport performance as it analyzes the problem from a global perspective and without considering the performance for each individual athlete. Therefore, in this work a heuristic (i.e. an approach to problem solving, that employs a practical method not guaranteed to be optimal, but sufficient) indicator is also proposed to analyze the model performance and its applicability to individual athletes from a practical perspective.

Concerning the applicability of the model in sport performance, there is a maximum error above which the model would not be useful. Defining as baseline that the precision error of the LT measurement is around 4-6\% \cite{Meyer1996}, we have enriched this information from the perspective of physiology in the following way: since higher level athletes require higher individualization in their daily trainings, a higher accuracy than the baseline is deemed necessary. On the other hand, this is the opposite for the less trained athletes which for the individualization is less critical \cite{Davison2009,Reilly2009}. The maximum error defined according to the race pace corresponding to the lactate threshold is shown in table \ref{tab:MaxError}.

\begin{table}
\centering
\begin{threeparttable}
\caption{Lactate threshold individualization error definition}
 \begin{tabular}{ccc} 
  \toprule
   Pace at LT	& \multicolumn{2}{c}{Maximum error in LT}\\
   (min/km) & $\pm$(s/km) & $\pm$(\%)\\
   \midrule
   $\ge$3\textsuperscript{$\ast$} & 3 & 1.7\\
   $\ge$3.5 & 5 & 2.4\\
   $\ge$4 & 10 & 4.2\\
   $\ge$4.5 & 15 & 5.5\\
   $\ge$5 & 20 & 6.6\\ 
   \bottomrule
  \end{tabular}
  \begin{tablenotes}
   \small
   \item *Out of scope: Fitness level above objective population
	 \item Abbreviations: LT, lactate threshold
  \end{tablenotes}
   \label{tab:MaxError}
 \end{threeparttable}
\end{table}

Therefore, the estimation error of each individual athlete is the difference between the real lactate threshold and the estimated lactate threshold. The model performance is then determined with the number of estimation errors below the maximum error defined in table \ref{tab:MaxError} and represented as the \% of the total athletes in the database (a.k.a. \% individualization). This value indeed represents the individualization power of the model (i.e. to percentage of correctly estimated lactate thresholds for the source population). 

Both performance indicators have their strengths: The statistical method gives insight about the global characteristics of the model while the heuristic does it about the individualization power of the model. Both indicators also allow to compare the performances between the training and test set which is essential to estimate the generalization power of the model. Therefore, the overall model performance is assessed and the models are ranked according to the best trade-off between both indicators and thus providing a solution to bridge the gap between theory and practice.

\section{Experimental setup}
\label{sec:experimental}

\subsection{Participants}
\begin{figure}
\includegraphics[width=\columnwidth]{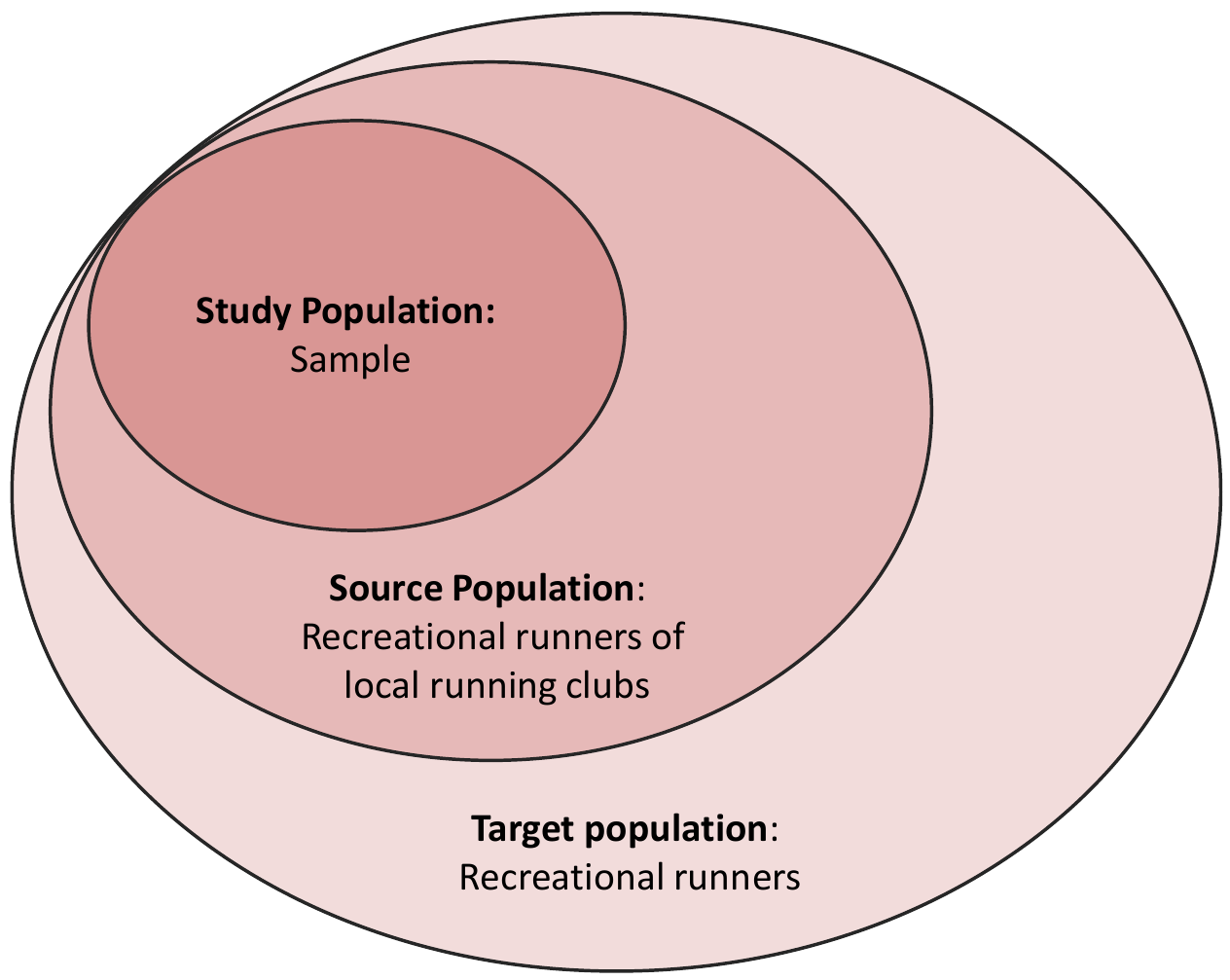}
\caption{Study population} 
\label{fig:Samp}
\end{figure}

The study included 143 recreational runners. The participants were selected from local running, triathlon and trail clubs so that the study population could be as representative as possible to the target population as represented in figure \ref{fig:Samp}. In this regard, several characteristics such as gender, age, discipline and level were aimed when selecting the athletes. In addition, personal interviews were also held so that other characteristics such us training background and current fitness level could be also assessed and thus a wide range of level athletes could be captured in the database.

Before participation, subjects were asked to present a medical certificate to ensure that they were free of cardiovascular, musculoskeletal and metabolic disease. This study was approved by the Ethics Committee for Research on Human Subjects of the University of Basque Country UPV/EHU (CEISH/GIEB). The athletes were informed about the tests and possible risks involved. Each athlete provided a written informed consent before testing.
 
All participants were federated in their respective disciplines, currently running at least 3 days a week and competing in recreational endurance races, and they had a running experience of at least 1 year. Athletes were encouraged to be well rested and to abstain from hard training sessions and competition for 24 hours before testing. All participants were familiarized with running on a treadmill.

\subsection{Data Acquisition Protocol}
All runners completed a maximal incremental running test at 1\% slope on a treadmill, which started at 9 km/h without previous warm up. The speed was increased by 1,5 km/h every 4 minutes until 13,5 km/h; then, to calculate the thresholds more accurately, speed was increased by 1 km/h until participant reached exhaustion. The duration of each step has been set in 4 minutes as it is adequate to determine the lactate threshold of endurance runners \cite{Santos-Concejero2014a}.

One minute of recovery was given between stages, when capillary blood samples were obtained from the earlobe and lactate concentration was measured by a portable lactate analyzer (Lactate Pro, Arkray, KDK Corporation, Kyoto, Japan.), which has been validated as an effective analyzer for lactate measurements \cite{Tanner2010}. In addition, respiratory RPE \cite{Aliverti2011} and muscular RPE \cite{Borg2010} was assessed using the 10-point Borg scale \cite{Borg1982a}. During the test, HR was monitored by an HR monitor (Garmin 910XT, Canton of Schaffhausen, Switzerland), measuring HRs just at the end of each stage and at the end of the 1 minute recovery. This recovery time is used because it provides better capacity to detect meaningful differences than HR measured after 2 minutes recovery \cite{Daanen2012}. 

Each test data was recorded both together with the incidences (if any) observed during the test, both in paper and electronic format. In addition, the HR data was downloaded from the HR monitor into a PC for its posterior analysis.

\subsection{Data pre-processing}
In order to detect incomplete or corrupted experimental data, the raw data recorded during the tests was analyzed. Three characteristics were sought: the compatibility between the HR monitor and manual data, the lack of incidences during the experiments which could distort the experimental data and a minimum of 14,5 km/h reached during the test. The minimum PTS required for introducing athletes into the model was set in 14,5 km/h because a minimum of 5 lactate measurement points are necessary to accurately determine the individual lactate threshold with Dmax method \cite{Santos-Concejero2014a}.  

The outlier experiments detected which could not be explained by incidences during the tests were kept in the database. 

After all the data was analyzed 105 test remained valid and were all gathered in a single database.

\section{Results and discussion}
\label{sec:results}
\subsection{Model fitting}
\subsubsection{Training configuration}
Table \ref{tab:PreRes} shows the steps followed in the preliminary training analysis to define the range of training algorithm configuration parameters to be used in the model fitting process. These preliminary trainings included data of 14 athletes that completed the 17,5 km/h stage. These athletes were selected as they are medium level recreational athletes which presumably have characteristics of the high and low level athletes. The configuration parameters have been tuned in three steps as explained in section \ref{sec:strategy}.

\begin{table}
\centering
\begin{threeparttable}
\caption{Preliminary trainings - training algorithm configuration parameters range definition}
 \begin{tabular}{@{}lll@{}} 
  \toprule
   Step & Hidden Units & Delay \\
  \midrule
   1 Coarse tuning & [1, 5, 10]  & [1, 3, 5, 8, 10]\\
   2 Increased resolution & [1, 2, 3, 4]  & [1, 2, 3, 4, 5, 6, 7, 8, 9, 10]\\
   3 Fine tuning & [1, 2, 3, 4]  & [5, 6, 7, 8, 9, 10, 11]\\
  \bottomrule																					
 \end{tabular}
\label{tab:PreRes}
\end{threeparttable}
\end{table}

Therefore, the ranges of training algorithm configuration parameters to be used in the model fitting are:
\begin{itemize}
\item 1-4  HUs
\item 5-11 Delays
\end{itemize}

\subsubsection{Train-test set selection}
As mentioned in section \ref{sec:strategy}, the modelling process is done comparing two different training and test set example selection methods. 

\begin{figure}
\includegraphics[width=\columnwidth]{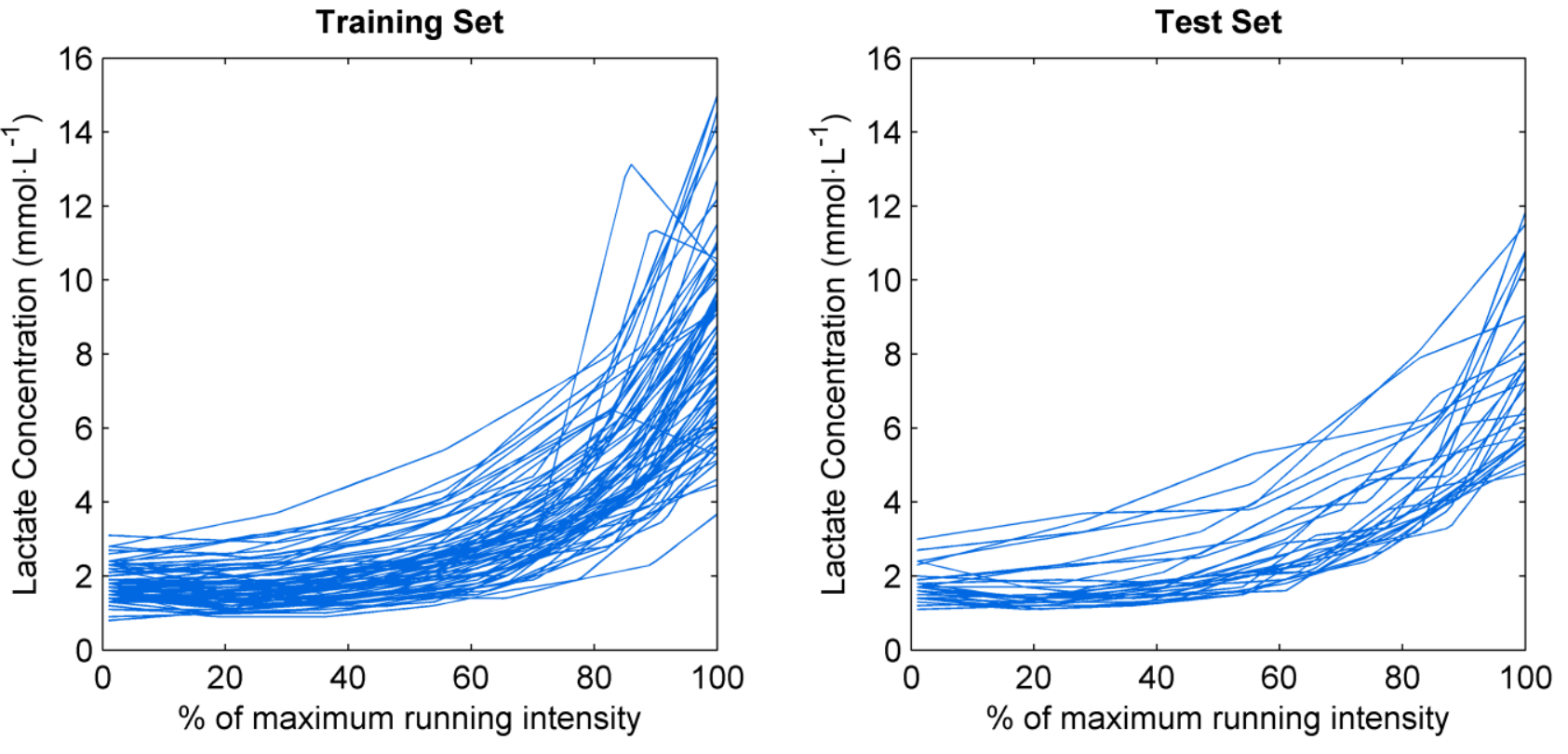}
\caption{Knowlege based sampling}
\label{fig:diferentShapes}
\end{figure}

On the one hand, figure \ref{fig:diferentShapes}, shows the test and training set spliting based on expert knowledge. 

\begin{figure}
\includegraphics[width=\columnwidth]{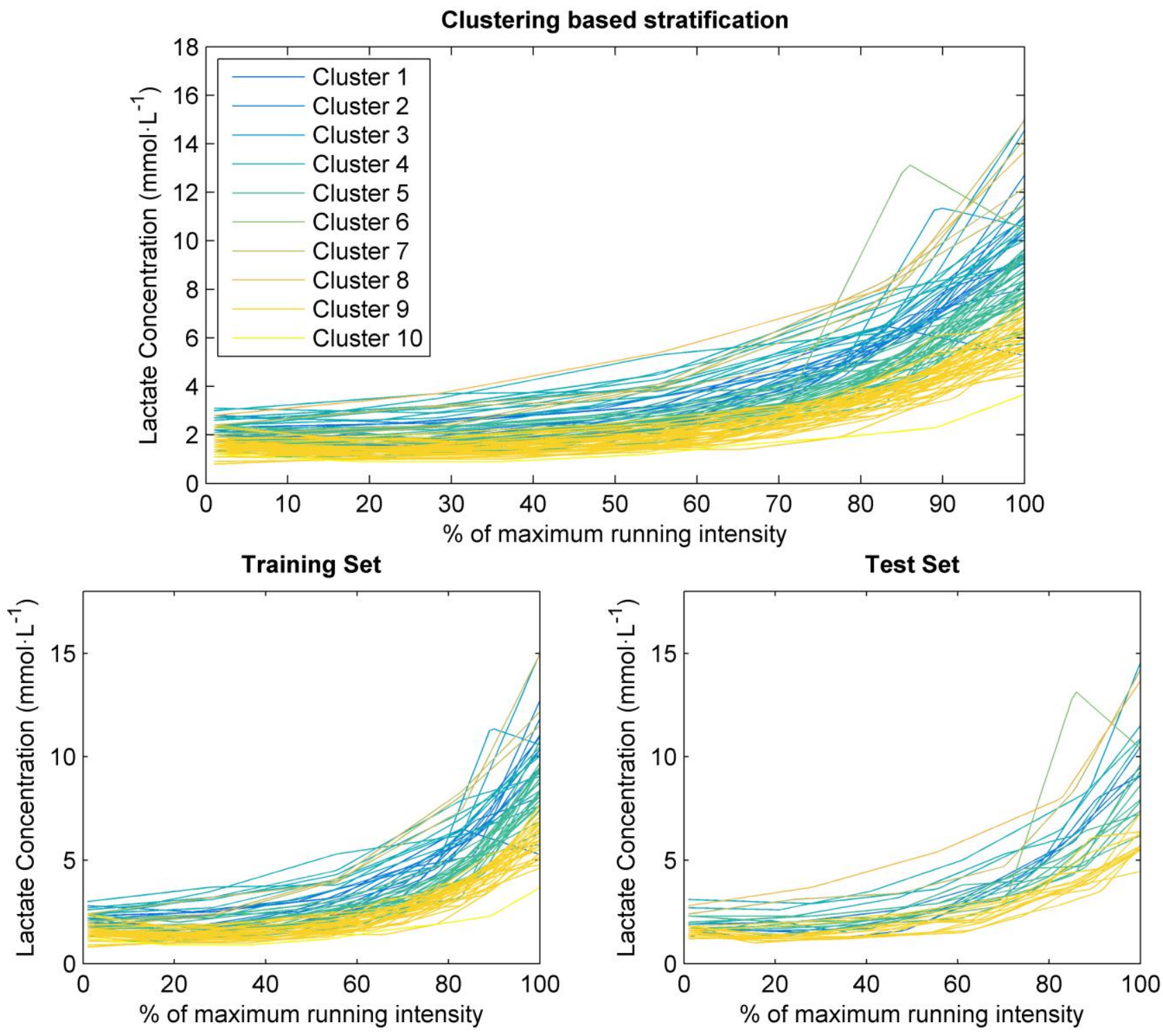}
\caption{Modified stratified sampling}
\label{fig:stratifiedSampling}
\end{figure}

On the other hand, figure \ref{fig:stratifiedSampling}, shows the training and test set selection using the modified stratified sampling method. Using the hierarchical clustering technique, 10 different stratum are found in the database. Then, the test set examples are selected random sampling the 30\% of the each stratum examples. The remaining examples correspond to the training set.

\subsection{Feature selection}
Using the different training configuration parameters and training and test set splits, multiple models are created and their performance analyzed according to the two performance indicators explained in section \ref{sec:strategy}. In addition, the training process is repeated for different input feature combinations following previously mentioned constructive criteria.

\begin{table}
\centering
\begin{threeparttable}
\caption{Knowledge based training-test set models performance respect to input features}
 \begin{tabular}{@{}ccc@{}} 
  \toprule
	 Input features	& Heuristic ind. & Statistic ind.\\
   	& \% indiv. & R (Test-Est LT)\\
	\midrule
   - & 90.48 & 0.89\\
   HRevo & 91.43 & 0.89\\
   HRevo, HRRevo & 91.43 & 0.89\\
   HRevo, HRRevo, RPEevo & 91.43 & 0.89\\
  \bottomrule
 \end{tabular}
 \begin{tablenotes}
  \small
  \item Abbreviations: \% indiv., \% individualization power or successfully estimated lactate thresholds; R(Test-Est LT), Pearson’s correlation coefficient between tested lactate threshold and estimated lactate threshold; HRevo, evolution of heart rate measured at the end of each stage; HRR, evolution of heart rate measured after 1 minute rest; RPEevo, evolution of the rate of perceived exertion measured at the end of each stage 
 \end{tablenotes} 
\label{tab:KBFeature}
\end{threeparttable}
\end{table}

\begin{table}
\centering
\begin{threeparttable}
 \caption{Modified stratified random sampling based training-test set models feature selection}
 \begin{tabular}{@{}ccc@{}} 
  \toprule
	 Input features	& Heuristic ind. & Statistic ind.\\
    & \% indiv. & R (Test-Est LT)\\
  \midrule
   - & 90.48 & 0.89\\
   HRevo & 91.43 & 0.89\\
   HRevo, HRRevo & 91.43 & 0.89\\
   HRevo, HRRevo, RPEevo & 91.43 & 0.89\\
  \bottomrule
 \end{tabular}
 \begin{tablenotes}
  \small
  \item Abbreviations: \% indiv., \% individualization power or successfully estimated lactate thresholds; R(Test-Est LT), Pearson’s correlation coefficient between tested lactate threshold and estimated lactate threshold; HRevo, evolution of heart rate measured at the end of each stage; HRR, evolution of heart rate measured after 1 minute rest; RPEevo, evolution of the rate of perceived exertion measured at the end of each stage 
 \end{tablenotes}
\label{tab:SSFeature}
\end{threeparttable}
\end{table}

As it is shown in tables \ref{tab:KBFeature} and \ref{tab:SSFeature}, the performances of the models are very similar. The model with no inputs performs well, meaning that the majority of the information of the lactate evolution is extracted from its own dynamic, something that can be explained by the similar dynamic that the lactate curve shows after standardizing it. In addition, introducing the HRevo increases the performance of the model due to the extra information that this feature provides. However, introducing HRRevo and RPEevo does not further improve the performance of the model which could mean that the information provided by these features is redundant and/or irrelevant.

Therefore, among the four combinations of input features, the model with one input (HRevo) is considered the best due to having the smallest error, as well as due to a lower complexity than the models with more input features.

\subsection{Model selection}
Once that the feature selection is done, a sensitivity analysis is performed in order to get insight about: \begin{itemize}
\item the relevance of the train-test set selection in the training process results.
\item the relation between the training configuration parameters (HU and Delays) and the performance of the model.
\end{itemize}

\begin{figure}
\includegraphics[width=\columnwidth]{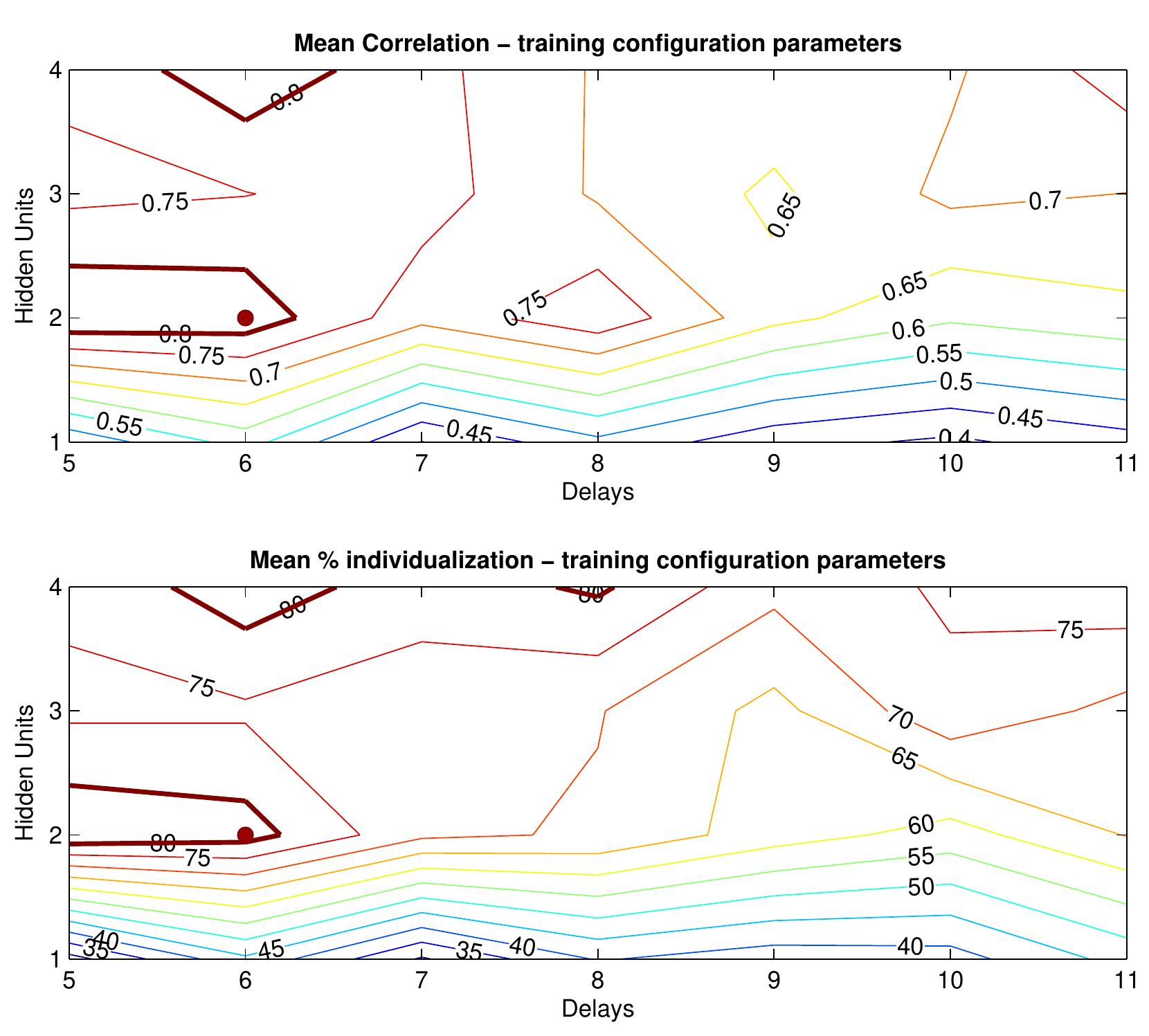}
\caption{Sensitivity analysis - Knowledge based sampling - configuration parameters VS performance}
\label{fig:KBPerfSA}
\end{figure}

\begin{figure}
\includegraphics[width=\columnwidth]{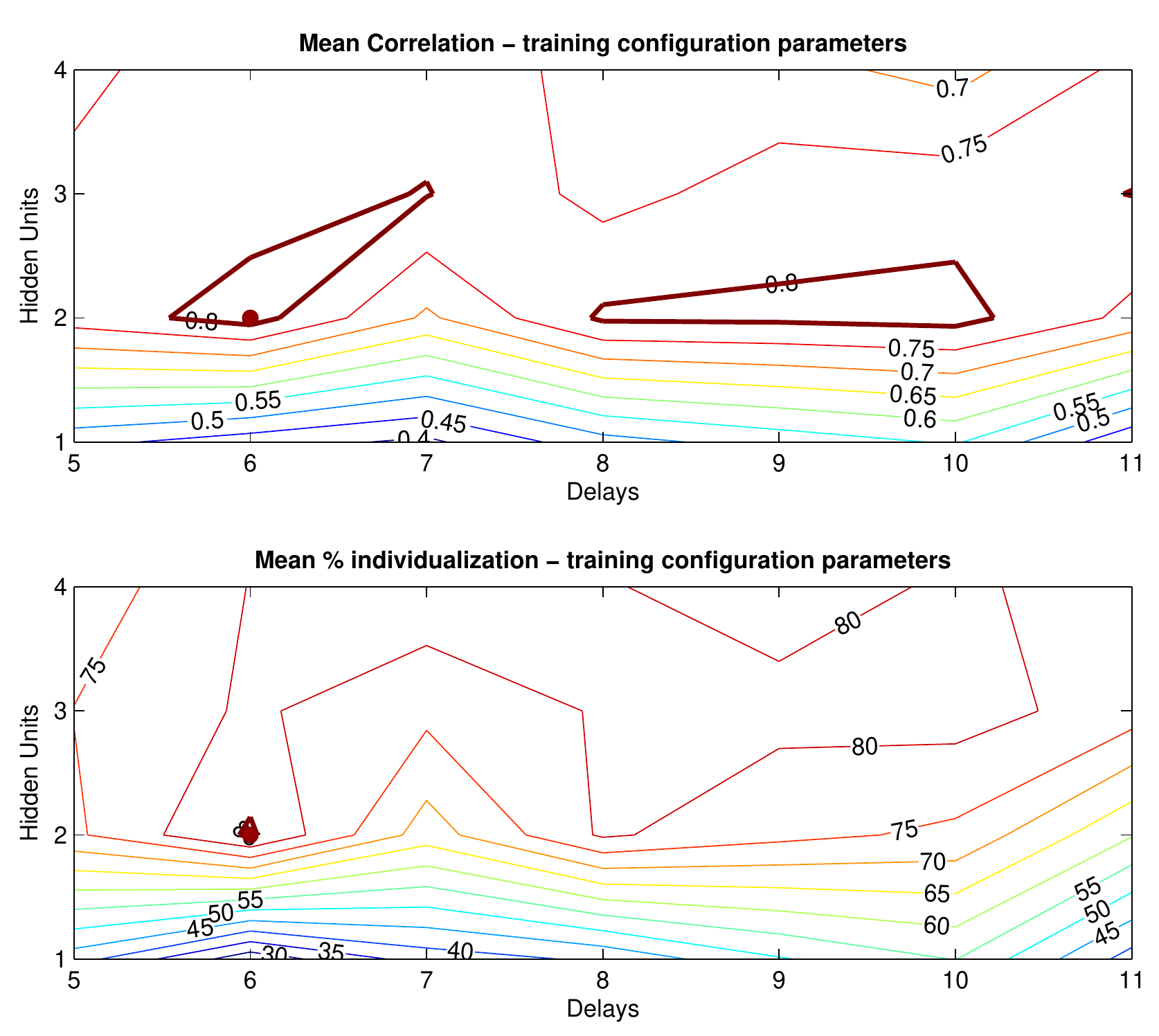}
\caption{Sensitivity analysis - Modified stratified random sampling - configuration parameters VS performance}
\label{fig:SSPerfSA}
\end{figure}

Figures \ref{fig:KBPerfSA} and \ref{fig:SSPerfSA} represent the performance of the models trained (according to both performance indicators) with respect to their configurations parameters for the knowledge based and the modified stratified sampling dataset splitting methods, respectively.  

On the one hand, the sensitivity analysis shows that the performances obtained with the modified stratified sampling technique are more homogeneous than the obtained with the knowledge based sampling method. This indicates that the models obtained with the modified stratified sampling can handle better the randomness of the data and thus presenting a more robust behavior. Therefore, the final model is selected within the models trained with this method.  

On the other hand, there are several configuration parameter zones that consistently obtain better performances across the analysis. Moreover, these zones are also consistent with both performance indicators thus validating the use of the heuristic performance indicator and the robustness of the proposed combined model performance estimation approach. 

From the generalization perspective the less complex zones are more interesting. Therefore, as highlighted in figures \ref{fig:KBPerfSA} and \ref{fig:SSPerfSA}, the best configuration parameter zone is consistently around 2 HU and 6 delays.

\subsubsection{Final model selection}
Finally, all the models trained with the modified stratified sampling method are ranked in decreasing performance order and the final model is selected according to the best parameter zone. Table \ref{tab:bestmodels} represents the 10 best models ranked according to the best trade-off between both model performance indicators.

\begin{table}
 \centering
 \begin{threeparttable}
 \caption{Best 10 model ranked by performance}
  \begin{tabular}{@{}cccccc@{}} 
  \toprule
	Rnk	& \multicolumn{2}{c}{Configuration} & Heuristic ind. & Statistic ind.\\
   & HU	& Del & \% indiv. & R (Test-Est LT)\\
  \midrule
  1.& 1 & 8 & 91.43 & 0.89\\ 
  2.& 1 & 7 & 91.43 & 0.89\\
  3.& 1 & 10 & 90.63 & 0.89\\
  4.& 4 & 11 & 90.63 & 0.89\\
  5.& 1 & 10 & 90.63 & 0.89\\
  6.& 2 & 6 & 89.52 & 0.89\\
  7.& 4 & 11 & 89.52 & 0.89\\
  8.& 2 & 8 & 89.52 & 0.89\\
  9.& 2 & 7 & 89.52 & 0.89\\
  10.& 3 & 10 & 89.52 & 0.89\\
  \bottomrule
  \end{tabular}
 \begin{tablenotes}
  \small
  \item Abbreviations: Rnk, Rank; HU, hidden units; Del, delays; \% indiv., \% of individualization power or successfully estimated lactate thresholds; R (Test-Est LT), Pearson’s correlation coefficient between tested lactate threshold and estimated lactate threshold.
 \end{tablenotes} 
\label{tab:bestmodels}
\end{threeparttable}
\end{table}

The final model selected is the one ranked in 6\textsuperscript{th} position as the performance is comparable to the best ones and is in the optimum configuration parameter zone identified in the sensitivity analysis.

\subsection{Applicability analysis: Individualization-Generalization power assessment}
In order to asses the applicability of the ML model here presented, the individualization and generalization power of the model are analyzed and gathered in table \ref{tab:FinalModel}. 

\begin{table}
\centering
 \begin{threeparttable}
 \caption{Final Model Performance}
  \begin{tabular}{@{}cccc@{}} 
  \toprule
   Perf. Ind. & Global Perf. & Train. Perf.	& Test. Perf.\\
   \midrule
   Heuristic ind. (\%) & 89.52 & 89.04 & 90.63 \\
   Statistic ind. (R) & 0.89 & 0.89 & 0.92 \\
   \bottomrule
  \end{tabular}
 \begin{tablenotes}
  \small
  \item Abbreviations: Perf, Performance; Ind., indicators; Train, Training set; Test, Test set; R, Pearson’s correlation coefficient between tested lactate threshold and estimated lactate threshold
 \end{tablenotes}	
 \label{tab:FinalModel}
 \end{threeparttable}
\end{table}

\begin{figure*}[!htbp]
\includegraphics[width=\textwidth]{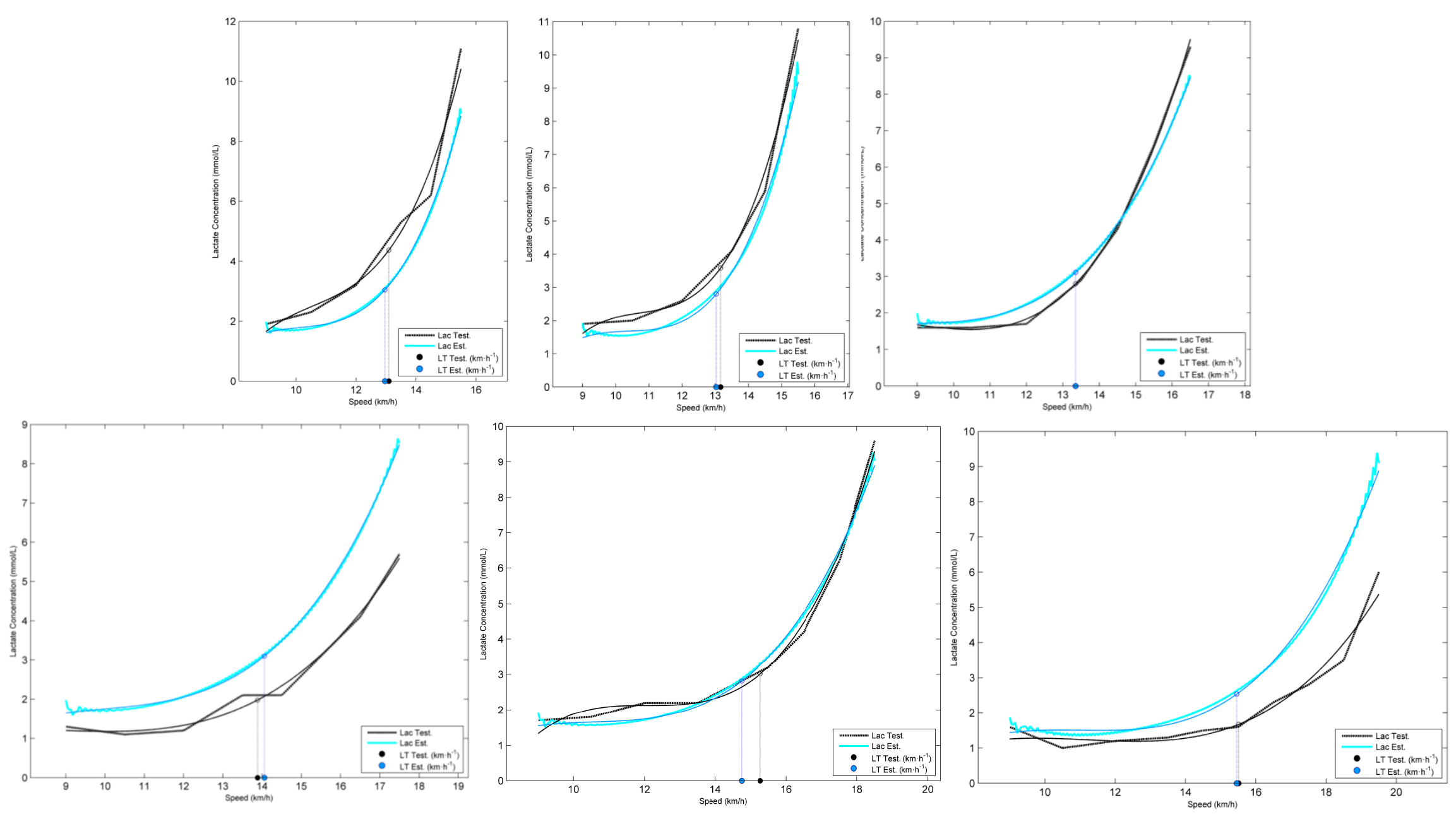}
\caption*{Abbreviations: Lac, Lactate; Test, tested; Est, estimated; LT, lactate threshold}
\caption{Correct lactate threshold estimations}
\label{fig:GoodEst}
\end{figure*}

On the one hand, the model estimates the lactate threshold of 89.52\% of the athletes with an error below the maximum acceptable error defined in table \ref{tab:MaxError}. This means that the model is capable of reaching the minimum individualization required in 89.52\% of the cases. Figure \ref{fig:GoodEst}, represents different correct lactate threshold estimations for different level athletes.

\begin{figure}[!htbp]
\includegraphics[width=\columnwidth]{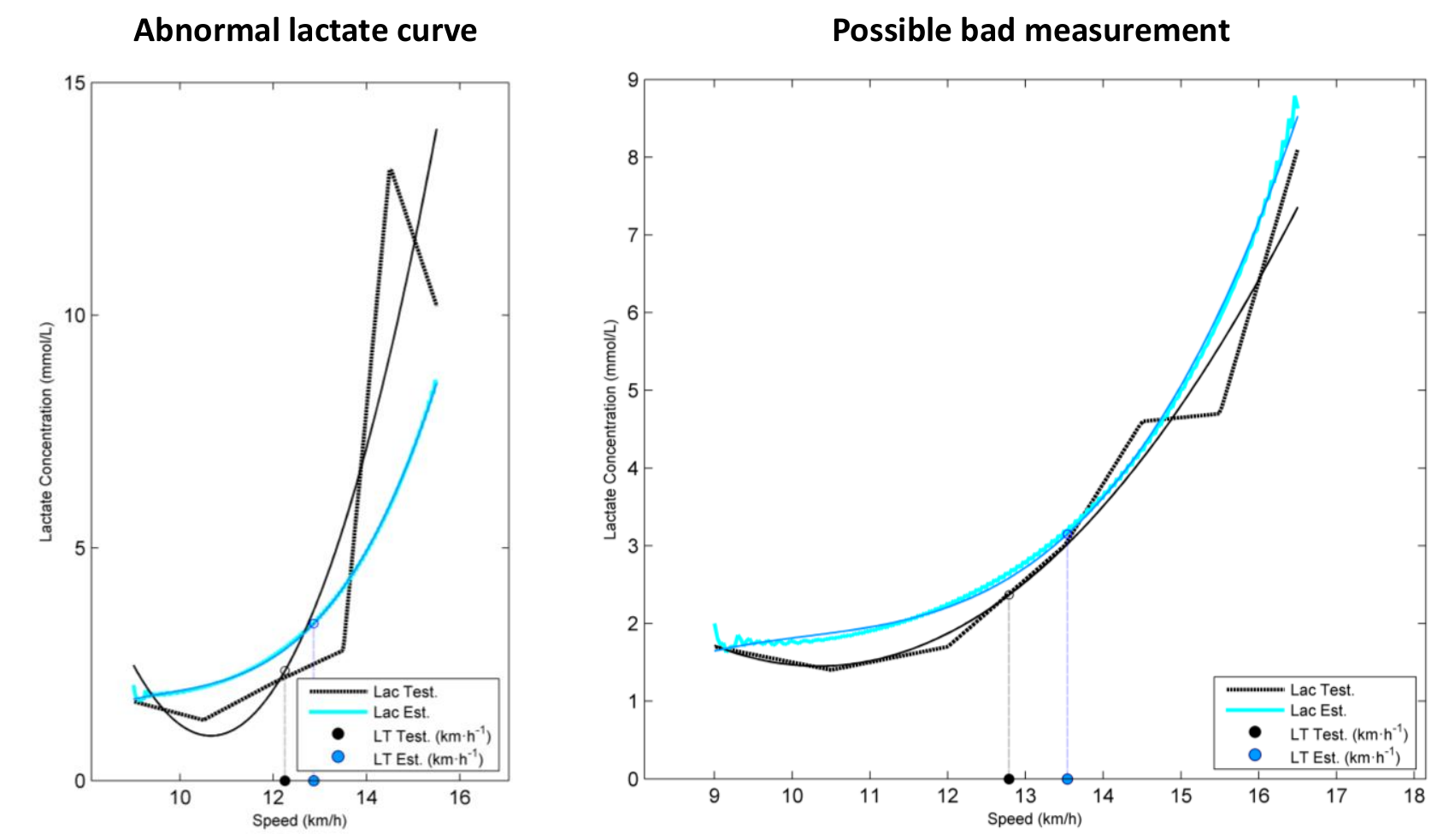}
\caption*{Abbreviations: Lac, Lactate; Test, tested; Est, estimated; LT, lactate threshold}
\caption{Bad lactate threshold estimations}
\label{fig:BadEst}
\end{figure}

Moreover, unsuccessfully estimated lactate thresholds coincide with abnormal lactate curves, which in some cases may be caused by bad blood lactate measurements as shown in figure \ref{fig:BadEst}.

On the other hand, the generalization power of the model is assessed comparing the percentage of correct estimations made in training and testing sets. As it is observed in table \ref{tab:FinalModel} both are very similar. This means that the generalization power of the model is very good, as the estimation of new athletes lactate threshold is as good as for the athletes used to train the model. 

These results also show that the final model is slightly shifted towards the bias end (as it is not able to fit the 100\% of the athletes into the model) of the bias-variance continuum. In any case, this is indeed the preferred solution when trying to attain generalization of complex systems as in our case.

In addition, in order to go deeper in this applicability analysis, a correlation analysis has been made between lactate threshold and athlete performance level. Its correlation with the athlete performance level is in fact one of the interesting characteristics of the individual lactate threshold applied in sports performance \cite{Pallares2016,Lacour1990}. In this work, the running performance is quantified with the best IAAF score obtained in certified races between 10 km and marathon distance during the last year \cite{B2011}. As it is shown in the table \ref{tab:PerfCorrelation}, the correlation of the estimated lactate threshold with the IAAF score increases comparing it to the lactate threshold calculated with the experimentally tested lactate values.

\begin{table}
\centering
 \begin{threeparttable}
 \caption{Correlation between lactate thresholds and running performance}
  \begin{tabular}{@{}ccc@{}} 
  \toprule
  - & R (IAAF score)\\
  \midrule
  Tested LT & 0.61 \\
  Estimated LT & 0.78 \\
  \bottomrule
  \end{tabular}
	\begin{tablenotes}
  \small
  \item Abbreviations: R, Pearson's correlation coefficient; LT, lactate threshold
 \end{tablenotes}	
  \label{tab:PerfCorrelation}
 \end{threeparttable}
\end{table}

This means that the individual lactate threshold estimated with the ML system is more correlated with the running performance than the experimentally tested one. Two reasons which could explain these phenomena are: 
\begin{itemize}
\item The estimated lactate curve is more homogenous as its not affected by blood lactate measurement errors which is known to be the main drawback of the blood lactate measurement \cite{Faude2009}. Consequently, the ML system acts as a filter and estimates the lactate threshold more accurately than the experimentally measured lactate threshold and increasing its correlation with the athlete performance indicators.
\item The ML system here proposed increases the collinearity between the lactate threshold and the PTS (from R=0.86 to R=0.97) due to a homogenization of the lactate curves estimated. As PTS is considered the best predictor of running performance \cite{Noakes1990} the estimated lactate threshold also correlates better with the IAAF score. Being this so, the increased correlation between the estimated lactate threshold and running performance would not mean that the estimated lactate threshold is more accurate than the experimentally tested lactate threshold, it would be a consequence of the methodology used.
\end{itemize}

Most probably both reasons are somehow involved in the improved correlation with the running performance. Nevertheless, this can not be quantitatively evaluated with the currently available information. Therefore, more tests would be needed to be able to conclude whether the ML system here proposed corrects some of the problems of the direct blood lactate measurement or its just a consequence of the ML methodology. 

In any case, it is clear that the correlation with the experimentally measured lactate threshold is very high (R=0.89) and that the correlation with the athlete performance is also high (R(IAAF)=0.78) meaning that the here proposed ML system is a good alternative to the traditional invasive lactate threshold measurement tests for recreational runners.

\section{Conclusions}
\label{sec:conclusions}

In this work, a ML system which estimates the individual lactate threshold of recreational athletes in a non-invasive, cost efficient and easily accessible way is presented. In addition to applying previously consolidated ML techniques to solve a complex real world problem, two novel methodologies have been proposed, the standardization of the temporal axis and a modification of the stratified sampling method. The ML system here presented is capable of successfully estimating the lactate threshold for 89.52\% of the study population and its generalization power with respect to the source population is very good as almost equal results have been obtained for the training and test sets. Estimating the generalization power on the target population is an extremely complicated task and dependent on the representativeness of the database and the ML methodology followed. Therefore, it's risky to assume that, despite making big efforts in the sampling process, the study population completely represents the target population. However, big efforts have been made from the methodological point of view to ensure that the final model's generalization power can be maximized and this is the main reason why a slightly underfitted model has been selected as the optimal. In addition, the correlation between the experimentally tested lactate threshold with the estimated lactate threshold (R=0.89) is very high. 

For all these reasons, the ML based system here presented is an alternative to the traditional invasive lactate threshold measurement tests for recreational runners. In the future it is planed to make further experiments to validate the ML system. In this regard two validations are foreseen: Against more new athletes and against athletes which are part of this database but which their fitness condition may have changed.  

\section*{Conflicts of interest}
The authors report no conflicts of interest.

\section*{Acknowledgements}
We thank to the members of the Department of Physiology and the Department of Physical Education and Sport for their support in the problem statement and during the tests. We also thank all the athletes for participation and Mikel Echezarreta for his help in the recruitment of the athletes and providing support during the tests.
Funding: This work was supported by Grupo Campus [projects LACTATUS 2016 and LACTATUS]; Department of Economic Development and Competitiveness of the Basque Government [Gaitek 2015]; University of the Basque Country UPV/EHU [projects PPG17/56 and PPG/17/40]; and Department of Education of the Basque Government [grant numbers IT914-16, US16/15 and PRE\_2015\_1\_0129].


\bibliography{ASC2017}

\begin{thebibliography}{10}
\expandafter\ifx\csname url\endcsname\relax
  \def\url#1{\texttt{#1}}\fi
\expandafter\ifx\csname urlprefix\endcsname\relax\def\urlprefix{URL }\fi
\expandafter\ifx\csname href\endcsname\relax
  \def\href#1#2{#2} \def\path#1{#1}\fi

\bibitem{Runningusa}
\href{http://www.runningusa.org/statistics}{{Running USA - cronological
  evolution of athletes participating in running events (accesed 26.04.2017)}}.
\newline\urlprefix\url{http://www.runningusa.org/statistics}

\bibitem{Triathlons}
\href{https://www.statista.com/statistics/191339/participants-in-triathlons-in-the-us-since-2006/}{{Triathlons
  USA - cronological evolution of athletes participating in thriatlon events
  (accesed 26.04.2017)}}.
\newline\urlprefix\url{https://www.statista.com/statistics/191339/participants-in-triathlons-in-the-us-since-2006/}

\bibitem{Vickers2016}
A.~J. Vickers, E.~A. Vertosick, {An empirical study of race times in
  recreational endurance runners}, BMC Sports Science, Medicine and
  Rehabilitation 8 (2016) 26.

\bibitem{Tanaka1984a}
K.~Tanaka, Y.~Matsuura, A.~Matsuzaka, K.~Hirakoba, S.~Kumagai, S.~O. Sun,
  K.~Asano, {A longitudinal assessment of anaerobic threshold and
  distance-running performance.} (jun 1984).

\bibitem{Pallares2016}
J.~G. Pallar{\'{e}}s, R.~Mor{\'{a}}n-Navarro, J.~F. Ortega, V.~E.
  Fern{\'{a}}ndez-El{\'{i}}aas, R.~Mora-Rodriguez, {Validity and reliability of
  ventilatory and blood lactate thresholds in well-trained cyclists}, PLoS ONE
  11 (2016) 1--16.

\bibitem{Lacour1990}
J.~R. Lacour, S.~Padilla-Magunacelaya, J.~C. Barth{\'{e}}l{\'{e}}my,
  D.~Dormois, {The energetics of middle-distance running.}, European journal of
  applied physiology and occupational physiology 60 (1990) 38--43.

\bibitem{Acevedo1989a}
E.~O. Acevedo, A.~H. Goldfarb, {Increased training intensity effects on plasma
  lactate, ventilatory threshold, and endurance}, Med.Sci.Sports Exerc. 21
  (1989) 563--568.

\bibitem{Hofmann2017}
P.~Hofmann, G.~Tschakert, {Intensity- and duration-based options to regulate
  endurance training}, Frontiers in Physiology 8~(MAY) (2017) 337.

\bibitem{Halson2014}
S.~L. Halson, {Monitoring Training Load to Understand Fatigue in Athletes} (nov
  2014).

\bibitem{Tanaka1984}
K.~Tanaka, Y.~Matsuura, {Marathon performance, anaerobic threshold, and onset
  of blood lactate accumulation.}, Journal of applied physiology: respiratory,
  environmental and exercise physiology 57~(3) (1984) 640--3.

\bibitem{Heck1985}
H.~Heck, A.~Mader, G.~Hess, S.~M{\"{u}}cke, R.~M{\"{u}}ller, W.~Hollmann,
  {Justification of the 4-mmol/l lactate threshold.}, International Journal of
  Sports Medicine 6~(3) (1985) 117--130.

\bibitem{Fay2009}
O.~Storen, J.~Helgerud, E.~M. Stoa, J.~Hoff, {Maximal Strength Training
  Improves Running Economy in Distance Runners}, Medicine and Science in Sports
  and Exercise 40~(6) (2008) 1087--1092.

\bibitem{Grant1997}
S.~Grant, I.~Craig, J.~Wilson, T.~Aitchison, {The relationship between 3 km
  running performance and selected physiological variables.}, Journal of Sports
  Sciences 15~(4) (1997) 403--410.

\bibitem{Santos-Concejero2014a}
J.~Santos-Concejero, R.~Tucker, C.~Granados, J.~Irazusta,
  I.~Bidaurrazaga-Letona, J.~Zabala-Lili, S.~M. Gil, {Influence of regression
  model and initial intensity of an incremental test on the relationship
  between the lactate threshold estimated by the maximal-deviation method and
  running performance.}, Journal of Sports Sciences 32 (2014) 853--9.

\bibitem{Peinado2016}
A.~Peinado, D.~{Pess{\^{o}}a Filho}, V.~D{\'{i}}az, P.~Benito,
  M.~{\'{A}}lvarez-S{\'{a}}nchez, A.~Zapico, F.~Calder{\'{o}}n, {The midpoint
  between ventilatory thresholds approaches maximal lactate steady state
  intensity in amateur cyclists}, Biology of Sport 33~(4) (2016) 373--380.

\bibitem{Conconi1982}
F.~Conconi, M.~Ferrari, P.~G. Ziglio, P.~Droghetti, L.~Codeca, {Determination
  of the anaerobic threshold by a noninvasive field test in runners}, Journal
  of Applied Physiology 52 (1982) 869--873.

\bibitem{Vachon1999}
J.~A. Vachon, J.~{Bassett, David R.}, S.~Clarke, {Validity of the heart rate
  deflection point as a predictor of lactate threshold during running}, J Appl
  Physiol 87~(1) (1999) 452--459.

\bibitem{Bourgois2004}
J.~Bourgois, P.~Coorevits, L.~Danneels, E.~Witvrouw, D.~Cambier, J.~Vrijens,
  {Validity of the heart rate deflection point as a predictor of lactate
  threshold concepts during cycling.}, Journal of strength and conditioning
  research 18~(3) (2004) 498--503.

\bibitem{Ringwood2014}
J.~V. Ringwood, J.~O. Neill, P.~Tallon, N.~Fleming, B.~Donne, {Non-invasive
  anaerobic threshold measurement using fuzzy model interpolation}, IEEE
  Conference on Control Applications (CCA) (2014) 1711--1715.

\bibitem{Borges2015}
N.~R. Borges, M.~W. Driller, {Wearable lactate threshold predicting device is
  valid and reliable in runners}, Journal of Strength and Conditioning Research
  30~(8) (2016) 2212--2218.

\bibitem{Proshin2013}
A.~P. Proshin, Y.~V. Solodyannikov, {Mathematical Modeling of Lactate
  Metabolism with Applications to Sports}, Automation and Remote Control 74~(6)
  (2013) 1004--1019.

\bibitem{Erdogan2009}
A.~Erdogan, C.~Cetin, H.~Goksu, R.~Guner, M.~L. Baydar, {Non-invasive detection
  of the anaerobic threshold by a neural network model of the heart rate–work
  rate relationship}, Proceedings of the Institution of Mechanical Engineers,
  Part P: Journal of Sports Engineering and Technology 223~(3) (2009) 109--115.

\bibitem{Borg1982a}
G.~Borg, {Psychophysical bases of perceived exertion.}, Medicine and science in
  sports and exercise 14~(5) (1982) 377--381.

\bibitem{LopezChicharro2004}
J.~{López Chicharro}, S.~{Aznar La{\'{i}}n}, A.~{Fern{\'{a}}ndez Vaquero},
  L.~M. {L{\'{o}}pez Mojares}, A.~{Luc{\'{i}}a Mulas}, M.~{P{\'{e}}rez Ruiz},
  {Transición aeróbica-anaeróbica: concepto, metodología de
  determinación y aplicaciones}, 1st Edition, Master Line {\&} Prodigio,
  2004.

\bibitem{Zhang2011}
N.~Zhang, {Prediction of Urban Stormwater Runoff in Chesapeake Bay Using Neural
  Networks}, in: Advances in Neural Networks – ISNN 2011, Springer, Berlin,
  Heidelberg, 2011, pp. 27--36.

\bibitem{Godarzi2014}
A.~A. Godarzi, R.~M. Amiri, A.~Talaei, T.~Jamasb, {Predicting oil price
  movements: A dynamic Artificial Neural Network approach}, Energy Policy 68
  (2014) 371--382.

\bibitem{Becerikli2007}
Y.~Becerikli, Y.~Oysal, {Modeling and prediction with a class of time delay
  dynamic neural networks}, Applied Soft Computing 7~(4) (2007) 1164--1169.

\bibitem{Babu2014}
C.~N. Babu, B.~E. Reddy, {A moving-average filter based hybrid ARIMA–ANN
  model for forecasting time series data}, Applied Soft Computing 23 (2014)
  27--38.

\bibitem{Laboissiere2015}
L.~A. Laboissiere, R.~A. Fernandes, G.~G. Lage, {Maximum and minimum stock
  price forecasting of Brazilian power distribution companies based on
  artificial neural networks}, Applied Soft Computing 35 (2015) 66--74.

\bibitem{Dreyfus2005}
G.~Dreyfus, {Neural Networks. Methodology and Applications}, Springer, Berlin,
  Heidelberg, 2005.

\bibitem{Asgari2016}
H.~Asgari, X.~Chen, M.~Morini, M.~Pinelli, R.~Sainudiin, P.~R. Spina,
  M.~Venturini, {NARX models for simulation of the start-up operation of a
  single-shaft gas turbine}, Applied Thermal Engineering 93 (2016) 368--376.

\bibitem{Teufel2003}
E.~Teufel, M.~Kletting, W.~G. Teich, H.~J. Pfleiderer, C.~Tar{\'{i}}n-Sauer,
  {Modelling the glucose metabolism with backpropagation through time trained
  elman nets}, in: Neural Networks for Signal Processing - Proceedings of the
  IEEE Workshop, Vol. 2003-Janua, IEEE, 2003, pp. 789--798.

\bibitem{Arriandiaga2015a}
A.~Arriandiaga, E.~Portillo, J.~a. S{\'{a}}nchez, I.~Cabanes, I.~Pombo, {A new
  approach for dynamic modelling of energy consumption in the grinding process
  using recurrent neural networks}, Neural Computing and Applications.

\bibitem{Matlab}
\href{https://es.mathworks.com/help/nnet/ref/layrecnet.html}{{Matlab NN- Layer
  recurrent neural network architecture (accesed 26.04.2017)}}.
\newline\urlprefix\url{https://es.mathworks.com/help/nnet/ref/layrecnet.html}

\bibitem{Hagan1994}
M.~T. Hagan, M.~B. Menhaj, {Training Feedforward Networks with the Marquardt
  Algorithm}, IEEE Transactions on Neural Networks 5~(6) (1994) 989--993.

\bibitem{MacKay1992}
D.~J.~C. MacKay, {A Practical Bayesian Framework for Backpropagation Networks},
  Neural Computation 4~(3) (1992) 448--472.

\bibitem{Mackay1995}
D.~Mackay, {Probable networks and plausible predictions — a review of
  practical Bayesian methods for supervised neural networks}, Network:
  Computation in Neural Systems 6 (1995) 469--505.

\bibitem{Nguyen1990}
D.~Nguyen, B.~Widrow, {Improving the learning speed of 2-layer neural networks
  by choosing initial values of the adaptive weights}, IJCNN Int. Joint Conf.
  Neural Networks 13 (1990) C21.

\bibitem{Ng2012}
A.~Ng, \href{http://cs229.stanford.edu/materials.html}{{Learning Theory - CS229
  - Lecture notes 4}}, Machine Learning (2012) 1--11.
\newline\urlprefix\url{http://cs229.stanford.edu/materials.html}

\bibitem{Machado2012}
F.~A. Machado, F.~Y. Nakamura, S.~M. F.~D. Moraes, {Influence of regression
  model and incremental test protocol on the relationship between lactate
  threshold using the maximal-deviation method and performance in female
  runners.}, Journal of sports sciences 30~(12) (2012) 1267--74.

\bibitem{Nicholson2001}
R.~M. Nicholson, G.~G. Sleivert, {Indices of lactate threshold and their
  relationship with 10-km running velocity.}, Medicine and science in sports
  and exercise 33~(2) (2001) 339--342.

\bibitem{Meyer1996}
K.~Meyer, {Ventilatory and lactate threshold determinations in healthy normals
  and cardiac patients: Methodological problems}, European Journal of Applied
  Physiology and Occupational Physiology 72~(5-6) (1996) 387--393.

\bibitem{Davison2009}
R.~C.~R. Davison, K.~A. van Someren, A.~M. Jones, {Physiological monitoring of
  the olympic athlete}, Journal of Sports Sciences 27~(13) (2009) 1433--1442.

\bibitem{Reilly2009}
T.~Reilly, T.~Morris, G.~Whyte, {The specificity of training prescription and
  physiological assessment: A review}, Journal of Sports Sciences 27~(6) (2009)
  575--589.

\bibitem{Tanner2010}
R.~K. Tanner, K.~L. Fuller, M.~L.~R. Ross, {Evaluation of three portable blood
  lactate analysers: Lactate Pro, Lactate Scout and Lactate Plus}, European
  Journal of Applied Physiology 109~(3) (2010) 551--559.

\bibitem{Aliverti2011}
A.~Aliverti, B.~Kayser, A.~L.~O. Mauro, M.~Quaranta, P.~Pompilio, R.~L.
  Dellac{\`{a}}, J.~Ora, L.~Biasco, L.~Cavalleri, L.~Pomidori, A.~Cogo,
  R.~Pellegrino, G.~Miserocchi, {Respiratory and leg muscles perceived exertion
  during exercise at altitude}, Respiratory Physiology and Neurobiology 177
  (2011) 162--168.

\bibitem{Borg2010}
E.~Borg, G.~Borg, K.~Larsson, M.~Letzter, B.~M. Sundblad, {An index for
  breathlessness and leg fatigue}, Scandinavian Journal of Medicine and Science
  in Sports 20 (2010) 644--650.

\bibitem{Daanen2012}
H.~A.~M. Daanen, R.~P. Lamberts, V.~L. Kallen, A.~Jin, N.~L.~U. {Van Meeteren},
  {A systematic review on heart-rate recovery to monitor changes in training
  status in athletes}, International Journal of Sports Physiology and
  Performance 7 (2012) 251--260.

\bibitem{B2011}
B.~Spiriev, {IAAF scoring tables 2011}, Monaco: Multiprint.

\bibitem{Faude2009}
O.~Faude, W.~Kindermann, T.~Meyer, {Lactate threshold concepts: How valid are
  they?} (2009).

\bibitem{Noakes1990}
T.~D. Noakes, K.~H. Myburgh, R.~Schall, {Peak treadmill running velocity during
  the VO2 max test predicts running performance.}, Journal of sports sciences
  8~(1) (1990) 35--45.

\end{thebibliography}

\end{document}